\documentclass{article}

\usepackage{PRIMEarxiv}

\usepackage[T1]{fontenc}    

\usepackage{amsmath,amssymb,amsfonts}
\usepackage{commath,bm}     
\usepackage{hyperref}
\usepackage{url}            

\usepackage{booktabs}       
\usepackage{nicefrac}       
\usepackage{subcaption}

\usepackage{xcolor}

\usepackage{graphicx}       
\graphicspath{{figures/}}     

\usepackage[linesnumbered,ruled]{algorithm2e}

\usepackage{multirow}
\usepackage{ctable} 
\usepackage{algpseudocode}
\usepackage{dsfont}
\usepackage{mathtools}
\usepackage{tablefootnote}

\usepackage{comment}

\SetKwInput{KwInput}{Input}                
\SetKwInput{KwOutput}{Output}              

\usepackage{tabularray}
\usepackage{xcolor,colortbl}

\definecolor{Gray}{gray}{0.85}
\definecolor{LightCyan}{rgb}{0.88,1,1}
\newcolumntype{g}{>{\columncolor{Gray}}c}
\definecolor{Gray2}{gray}{0.9}
\newcolumntype{G}{>{\columncolor{Gray2}\arraybackslash}c}

\usepackage{mathtools}
\DeclarePairedDelimiter{\ceil}{\lceil}{\rceil}

\def\BibTeX{{\rm B\kern-.05em{\sc i\kern-.025em b}\kern-.08em
    T\kern-.1667em\lower.7ex\hbox{E}\kern-.125emX}}

\title{Financial Data Analysis with Robust Federated Logistic Regression}

\author{
  \normalfont Kun Yang\thanks{\small This work was completed while the author was at Princeton University. Email: \texttt{kun88.yang@gmail.com}}
  \and
  \normalfont Nikhil Krishnan\thanks{\small Department of Operations Research and Financial Engineering, Princeton University, Princeton, NJ 08544, USA. Email: \texttt{nikhil.krishnan@princeton.edu}}
  \and
  \normalfont Sanjeev R. Kulkarni\thanks{\small Department of Electrical and Computer Engineering, Princeton University, Princeton, NJ 08544, USA. Email: \texttt{kulkarni@princeton.edu}}
}

\begin{document}
\maketitle




\begin{abstract}
In this study, we focus on the analysis of financial data in a federated setting, wherein data is distributed across multiple clients or locations, and the raw data never leaves the local devices. 
Our primary focus is not only on the development of efficient learning frameworks (for protecting user data privacy) in the field of federated learning but also on the importance of designing models that are easier to interpret. In addition, we care about the robustness of the framework to outliers.
To achieve these goals, we propose a robust federated logistic regression-based framework that strives to strike a balance between these goals.
To verify the feasibility of our proposed framework, we carefully evaluate its performance not only on independently identically distributed (IID) data but also on non-IID data, especially in scenarios involving outliers.
Extensive numerical results collected from multiple public datasets demonstrate that our proposed method can achieve comparable performance to those of classical centralized algorithms, such as Logistical Regression,  Decision Tree, and K-Nearest Neighbors, in both binary and multi-class classification tasks.
\end{abstract}




\section{Introduction}
\label{sec:introduction}
Financial data analysis plays a pivotal role in today's business landscape \cite{ahmed2022artificial, sahu2023overview, akomea2022review,  weber2024applications, goodell2021artificial, malhotra2024predicting, hojaji2022machine}, including credit risk assessment (such as loan prediction and credit scoring), fraud detection, and cost optimization, etc.
However, when we develop solutions to address financial problems, we will inevitably encounter a number of key challenges \cite{ahmed2022artificial, sahu2023overview, akomea2022review,  weber2024applications, goodell2021artificial}. For example, financial data is often voluminous, dynamically and frequently generated in real time, and distributed across diverse locations, making it challenging to process and analyze in a centralized manner\cite{ahmed2022artificial}, e.g., the New York Stock Exchange (NYSE) alone has billions of transactions per day. Similarly, other major exchanges, such as the Shanghai Stock Exchange (SSE) and the London Stock Exchange (LSE), also generate vast amounts of stock data. 
Additionally, noise and missing values unavoidably occur in financial data, which can cause results and predictions to be skewed (or even completely wrong). These challenges require firms to come up with more efficient and smarter solutions.  

In recent decades, machine learning has achieved remarkable success across various domains  \cite{sarker2021machine, xu2015comprehensive, min2018survey}, owing to its effective generalization ability and adaptability, and has also received increasing attention in financial data analysis \cite{ozbayoglu2020deep, cao2022ai}, such as credit risk assessment, resource allocation, and cost optimization. 
However, these classical (supervised) machine learning based solutions, such as logistic regression and random forest, usually implicitly assume that 
1) all the data is stored and centralized at one location, typically a single machine, and that we have full access to the entire data;
2) these algorithms expect to run on a single machine with minimal concerns for memory or disk storage limitations; and 
3) the provided data is clean and free from outliers introduced by malicious adversaries, as it is stored at a single location equipped with high security protection mechanisms to prevent data corruption. 
Nonetheless, these assumptions do not always hold in practice.
Federated learning \cite{DCKonecnyFederated} has been proposed to eliminate the first two assumptions, and it has already shown promising performance in various applications \cite{liu2022distributed, mothukuri2021survey}. 
Nonetheless, the underlying assumption of the requirement of clean data in the vanilla federated learning framework is still there. 
Another crucial aspect to consider in machine learning (and federated learning)-based frameworks revolves around interpretability and explainability of the solutions \cite{burkart2021survey}, 
which refers to the ease with which humans can understand and make sense of a machine learning model's predictions, decisions, and internal workings. 

Considering these issues and concerns, our goal is to effectively analyze financial data distributed across clients or locations using machine learning algorithms, while also ensuring that the model possesses strong interpretability and explainability, under some real concerns \cite{burkart2021survey},  such as safeguarding user privacy, addressing outliers introduced during data generation and transmission, countering malicious data generated by adversaries, managing data transmission costs, and complying with legal regulations governing data sharing and collection \cite{gdpr2023, hipaa2023}.
Thus, some complex algorithms (such as random forests, neural networks, and large language models, which can achieve high detection accuracy, but usually tend to exhibit lower levels of interpretability and explainability, and may consequently incur higher data transmission costs) are out of scope of this work.
To this end, we propose a federated logistic regression (FLR) based framework to efficiently address these challenges.  
We summarize the key aspects of our algorithm as follows:

\begin{itemize}
\item 
This approach only requires sending limited additional information (specifically, coefficients) from the clients to the server to learn a global model based on the collective information from all the clients (to manage transmission costs). It also avoids the need to gather and access raw data located or stored at the edges/clients (to protect user privacy). Furthermore, it has a very good interpretability as the coefficients of logistic regression directly demonstrate their significance and influence on the model. Thus, we can have more confidence to leverage the results generated by the model to make a (financial) decision. 

\item Our federated learning framework demonstrates  comparable AUC performance 
in both binary and multi-class classification tasks on four public datasets in Section \ref{sec:experiments}), when compared to the conventional centralized supervised machine learning algorithms, such as logistical regression, decision tree, and k-nearest neighbors. 

\item Moreover, by employing robust aggregation strategies (including coordinate-wise median and trimmed mean) within federated learning frameworks, our approach demonstrates consistent AUC performance across both IID and non-IID data, even when confronted with varying percentages of outliers. 

\end{itemize}

The remaining part of this paper is structured as follows. In Section \ref{sec:related_work}, we provide an overview of recent research concerning centralized and federated learning algorithms in finance. Moving on to Section \ref{sec:proposed_method}, we present our formulation and notation, and introduce our federated logistic regression framework tailored for finance data analysis. Section \ref{sec:evaluation_metrics} introduces classical supervised machine learning algorithms and the performance evaluation metrics employed in this study. In Section \ref{sec:experiments}, we conduct a comparison between our proposed algorithm and the centralized algorithms discussed in Section \ref{sec:evaluation_metrics}, utilizing various publicly available datasets. Following the experimental analysis in Section \ref{sec:experiments}, we discuss the limitation of this work in Section \ref{sec:discussion}. Subsequently, we conclude our work and outline our future directions in Section \ref{sec:conclusion}.

\section{Related work}
\label{sec:related_work}

In this section, we will briefly introduce the application of supervised machine learning algorithms in the field of financial data analysis, without delving into an exhaustive survey of machine learning algorithms in general, which can be found in survey works such as \cite{goodell2021artificial, bhatore2020machine, malhotra2024predicting, hojaji2022machine}. 

\textbf{Centralized Algorithms for Finance}: Financial uses of machine learning (\cite{dixon20} gives an introductory account) typically assume a centralized setting, such that all the raw data is located at a single node, e.g. a server, and that we have sufficient memory or disk storage available to process all the data on the server.  
Many standard machine learning approaches have been studied in a financial context, such as neural networks \cite{kurani23, coakley00}, support vector machines \cite{kurani23}, and reinforcement learning \cite{hambly23}.
Moscato et al. \cite{moscato2021benchmark} proposed a benchmarking analysis with the objective of evaluating the effectiveness of various credit risk scoring models commonly used for predicting if a loan will be repaid in a real social lending platform (Lending Club) dataset. 

\textbf{Federated Learning Algorithms for Finance}: 
Federated learning has been proposed to 
address the problems existed in the centralized frameworks, such as distributed data and limited memory \cite{jain2010data, agarwal2011issues, kant2009data, yang2019federated}.
Long et al. \cite{long2020federated} discussed the possible challenges of applying federated learning in the context of open banking. 
In particular, their paper focused on discussing the statistical heterogeneity, model heterogeneity, access limits, and one-class problems that are rarely discussed in other places. 
Imteaj et al. \cite{imteaj2022leveraging}  
proposed a federated learning framework for predicting customers’ financial distress while simultaneously addressing privacy concerns and straggler issues. 
Lee et al. \cite{lee2023federated} introduced a federated learning prototype featuring four layers of long short-term memory. This innovation enabled smaller financial institutions to enhance their competitiveness by engaging in collaborative machine learning model training. 
Moreover, Shingi et al. \cite{shingi2020federated} proposed a federated learning-based approach with neural networks for predicting loan applications that are unlikely to be repaid by sharing model weights aggregated at a central server.
Notably, their experiment lacked outliers in the dataset.

\textbf{Model Interpretability:}
In finance, we need models that not only perform well in terms of accuracy/AUC, but also excel in interpretability.
Bella et al. \cite{belle2021principles} provided a comprehensive introduction to the numerous advancements and facets within the realm of explainable machine learning. 
Burkart  et al. \cite{burkart2021survey} provided a detailed introduction to the topic of explainable supervised machine learning regarding the definitions and a foundation for classifying the various approaches in the field.
Additionally, Blanco et al. \cite{blanco2019machine} focused on explaining black-box models by using decision trees of limited size as a surrogate model. 

While numerous algorithms have been proposed for financial data analysis, most of them primarily focus on achieving high detection performance in centralized settings. Although several distributed learning frameworks have already been introduced to handle data distributed across multiple clients or locations, the proposed solutions often rely on advanced machine learning algorithms like neural networks to achieve their expected performance, which can impede the interpretability of their model outputs.
In this paper, we emphasis on not only developing an efficient learning framework in the context of federated learning but also underscore the importance of designing more easily interpretable models.

\section{Proposed Method}
\label{sec:proposed_method}
\subsection{Notation}
Before we delve into our proposal, it's necessary to introduce our notation. In this work, we consider an environment with $M$ clients, where each client $m$ has $n_m$ data points represented by $\mathcal{X}_m$. The entire dataset across the $M$ clients contains ${N = \sum_{m=1}^{M} n_m}$ data points represented by ${\mathcal{X} = \bigcup_{m=1}^{M} \mathcal{X}_m}$.
$\lfloor \cdot \rfloor$ is round down a given float to integer. 
Furthermore, at time (or iteration) $t$, $W_m(t) \in \mathbb{R}^d$ denotes the parameters of the local model for the client $m$, and $W(t) \in \mathbb{R}^d$ denotes the parameters of the global model at the server obtained based on $W_m(t-1)$ from all clients (or a subset of clients). During the learning phase, $\eta_m$ denotes the learning rate used on the client $m$, and $\mathbf{\sigma(\cdot)}$ denotes the sigmoid function used in the logistic regression model.
Additionally, $C$ denotes the number of classes in classification tasks, and $[M]$ denotes the set of integers $\{1, 2, \ldots, M\}$. 

\subsection{Problem Formulation}
In this work, we consider a distributed learning framework where our goal is to uncover the potential or underlying relationships between the features and the target class. Specifically, 
we aim to ascertain the significance of each feature concerning its dependency on the target variable while the data remains distributed across the $M$ clients, with the help of a central server.  

\subsubsection{Binary Federated Logistic Regression (BFLR)}
\label{Sec:flr}
We first concentrate on the binary classification task, where the target class possesses only two distinct values. This task is the simplest encountered in practical scenarios, such as predicting whether a client's loan will become problematic in the near future or not. In the subsequent section, we will introduce how to extend the binary logistic regression model onto a multi-class classification model.

Our goal of this section is to learn a global logistic regression model under the federated learning framework. 
Due to the characteristics of the target variable $Y$, which only takes two discrete values in the binary classification task with $Y=1$ representing that an event occurs (and it's a positive class), while $Y = 0$ representing the event doesn't occur (and it's a negative class),  
classical linear regression does not seem to be a appropriate model for this scenario because it assumes a linear relationship between $X \in \mathbb{R}^{d \times N}$ and $Y \in \mathbb{R}^{N}$; however, it is not the case here. Instead, we adapt logistic regression that assumes a linear relationship exists between the logistic (logit) transformation of $Y$ and $X$ \cite{maalouf2011logistic}, 
\begin{equation}
    logit(Y) = X^TW + b \label{equ:logit}
\end{equation}
i.e., our aim is to predict the logit of $Y$ based on $X$, rather than directly predicting $Y$ from $X$. Here, $W \in \mathbb{R}^d$ are coefficients and $b \in \mathbb{R}$ is the intercept of the logistic regression model. 
It is worth noting that the logit function can also be expressed as follows:
\begin{equation}
    logit(Y) = \ln(odds) = \ln(\frac{p}{1-p}), \label{equ:odds}
\end{equation}
where odds is a ratio of probabilities $p$ of $Y$ happening to probabilities ($1-p$) of $Y$ not happening, and $p$ is the probability of $Y$ being 1 given the input $X$, i.e., $P(Y=1|X)$. 
Combining (\ref{equ:logit}) and (\ref{equ:odds}) we have 
\begin{equation}
    p = \frac{1}{(1 + e^{X^TW+b})} = \sigma(-({X}^T{W} + b)),
\end{equation}
where $\sigma(\cdot)$ is the logistic cumulative distribution function (CDF), which maps real values ($-\infty$, $\infty$) to the unit interval [0, 1].


To estimate the parameters $(W, b)$, we use maximum likelihood estimation (MLE) \cite{czepiel2002maximum}, one of the most commonly used parameter estimation methods. 
With the assumption that the observed data points are independent, we formulate the likelihood function as follows:
\begin{equation}
    \mathcal{L} = \prod_{i=1}^{N} p^{Y_i} (1-p)^{1-Y_i},
\end{equation}
and hence, the log-likelihood is then,
\begin{align}
    & \ln(\mathcal{L}) = \ln(\prod_{i=1}^{N} p^{Y_i}  (1-p)^{1-Y_i})  
     =  \sum_{i=1}^{N} {Y_i} p + (1-Y_i)(1-p).
\end{align}
Then we can formulate the problem  as 
\begin{align}
    \underset{W, b}{\text{argmin}}\ \ln(\mathcal{L}) =  \underset{W, b}{\text{argmin}}\sum_1^{N} (Y_i p + (1-Y_i)(1-p)),
    \label{equ:argmin_lnL}
\end{align}
i.e., we would like to find a pair of $(W, b)$ that can minimize the cost function (\ref{equ:argmin_lnL}).

Our algorithm is outlined in Algorithm \ref{alg:flr}. At a high-level perspective, it consists of three main components: initialization, server update, and client update.

\textbf{Initialization}: We first instantiate a global logistic regression model with default parameters, along with individual local model for each client, also initialized with default parameters. 
For simplicity, at the server, we initialize the parameters $(W$ and $b)$ to 0 and then broadcast them to all clients. 
The clients, in turn, utilize the received global parameters to initialize their respective local models. 
Once the initialization is completed, we can proceed to the next iteration.
It's worth noting that by employing more sophisticated initialization techniques \cite{sutskever2013importance}, the model may converge faster or achieve better performance in terms of accuracy or AUC. Nevertheless, it is beyond the scope of this study. 
%
These processes correspond to the lines 6 to 9 in Alg. \ref{alg:flr} at iteration $t=0$.

\textbf{Server Update}: 
\label{server_update}
During each iteration ($t>0$ after the initialization), the server first broadcasts the global parameters $(W, b)$ to the clients, and then waits for the updated parameters sent from the clients learned from their local data. 
Once the server receives the updated parameters $(W_m(t), b_m(t))$, where $m \in [M]$ (which is the index of the random selected client at time $t$) it will aggregate them to obtain the new global parameters $(W(t), b(t))$. 
We would like to emphasize that the performance of the model heavily relies on the aggregation strategies we employ. Undoubtedly, the most straightforward aggregation method used is to simple take the \textbf{mean} of the parameters (i.e., coefficients $W$ and intercepts $b$) for each coordinate of the algorithm collected from the clients to obtain aggregated parameters.
At $t$-th iteration, the coefficients $W(t)$ and intercepts $b(t)$ can be computed as follows:
\begin{align}
   W(t) &= \frac{1}{M} \sum_{m=1}^{M} W_m(t), \nonumber \\ 
   b(t) &= \frac{1}{M} \sum_{m=1}^{M} b_m(t), \nonumber
\end{align}
where $W_m(t)$ and $b_m(t)$ are the local parameters of the $m$-th client.
Unless otherwise specified, we simply refer to this aggregation strategy as the default aggregation strategy for FLR, denoted as \textbf{FLR-mean}.
However, it is very sensitive to outliers (even a single outlier can cause it to fail). To mitigate the effects of outliers and missing values, robust estimators are preferable. 
In this paper, we do not aim to exhaust all the robust methods/strategies. Instead, our primary objective is to demonstrate that with robust aggregation methods, we can effectively alleviate the impact of outliers.
Consequently, within the scope of this study, we  concentrate on two widely adopted robust aggregation methods: coordinate-wise median and trimmed mean.
Instead of averaging the parameters (i.e., coefficients and intercepts) collected from the clients to obtain aggregated parameters, we choose the median values of the parameters of each coordinate with $median(\cdot)$ to obtain the aggregated parameters. 
Note that there are other aggregation methods \cite{tukey1977exploratory}; however, they may be too complicated to implement in practice.
As before, we compute $W(t)$ and $b(t)$ as follows:
\begin{align}
    W(t) &= median (W_m(t)), \nonumber \\ 
    b(t) &= median (b_m(t)), \nonumber
\end{align}
We refer to this kind of aggregation strategy as FLR with median aggregation, denoted as \textbf{FLR-median}.
On the other hand, we trim the top and bottom some percent $\alpha$ of data for each coordinate, and then average the rest of data to obtain the final aggregated parameters by $trimmed\_mean(\cdot)$. 
As before, we compute $W(t)$ and $b(t)$ as follows:
\begin{align}
    W(t) &= trimmed\_mean (W_m(t), \alpha), \nonumber \\ 
    b(t) &= trimmed\_mean (b_m(t), \alpha), \nonumber
\end{align}
We refer to this kind of aggregation strategy as FLR with trimmed mean aggregation, denoted as \textbf{FLR-trim\_mean}. This method requires some value for trimmed percentage $\alpha$ of data, which should generally be at least half of the percentage of outliers $p_{out}$, i.e. $\alpha \geq p_{out}/2$.

Then the current iteration finishes, and the algorithm proceeds to the next iteration.
All these processes correspond to the lines 11 to 12, and lines 15 to 17 
at iteration $t$ in Alg. \ref{alg:flr}. 

\textbf{Client Update}: During the same iteration $t$, the clients update their local parameters $(W_m(t), b_m(t))$ based on the global parameters $(W(t-1), b(t-1))$ received from the server 
\begin{align}
    W_m(t) &= W(t-1) - \eta \nabla_{W} \mathcal{L}, \nonumber \\ 
    b_m(t) & = b(t-1) - \eta \nabla_{b} \mathcal{L}, \nonumber
\end{align}
where $\eta$ is the learning rate, $\nabla$ is the partial derivative of the cost function regarding $W$ and $b$, and the loss function is cross-entropy loss. 
Note that $\eta$ can be different on different clients, and can be adapted over time \cite{liu2019variance}. However, for simplicity, we use the same and fixed $\eta$ in this work.

Clients can update their local parameters either once or multiple times using their own data. To reduce the overall number of the iterations between the server and clients, we update the local clients parameters multiple times, but only update the server's parameter once. 
Subsequently, we send the updated parameters $(W_m(t), b_m(t))$ back to the server and then wait for the next iteration.  Also, to mitigate overfitting in the local model, we apply the $\ell_2$ penalty on the loss function.  
Moreover, among the clients, there may be some clients that are compromised by adversaries (due to weak security protection mechanisms), or whose local data has very noisy data (due to the data generation and measurement process; for example some clients may have noisy data due to their misconduction/misoperations). Either of these will cause the model to output very biased local parameters, which will further impact the aggregation of the global parameters at the server for the vanilla federate learning framework with mean aggregation. 
We will verify this on Section \ref{sec:experiments}, and the results will show that robust aggregation methods can still work under the existence of outliers during the learning phase. More detail can be seen in Section \ref{sec:experiments}. 
All of these steps correspond to lines 13 to 14 at iteration $t$ in Alg. \ref{alg:flr}.

Using the methods presented in \textbf{Server Update} and \textbf{Client Update}, we iteratively update the parameters of both the server and clients until the global model converges. In Alg \ref{alg:flr}, when the number of iterations reaches to the maximum number of iterations $T$ preset by our before the learning task, the learning phase stops.

\begin{algorithm}
\caption{Federated Logistic Regression (FLR)}
\label{alg:flr}
\KwInput{ $T$ (maximum number of iterations); $M$ (number of clients);
$\gamma$ (ratio of clients selected at each training iteration); $T_c$ (number of iterations allocated for local parameter updates on each client); and $agg\_method$ (the method employed for aggregating parameters from the clients).}
\KwOutput{Final parameters (${W}$, $b$).}

\For{each iteration $t=0, \ldots, T$}{
     \# \texttt{(1) Randomly select clients for training the model (Optional).} \\ 
    \uIf{$t\%10 == 0$}{
        $M_t \leftarrow$  Randomly select $\ceil{\gamma \cdot M}$ clients. \\ 
    }
    \# \texttt{(2) Initialize the parameters of FLR.} \\ 
    \uIf{t == 0}{
         $\mu$, $\sigma$ $\leftarrow$ compute all clients' $\mu$s and $\sigma$s. \\ 
         Standardize each client data with ($\mu$, $\sigma$). \\ 
        ${W} (t) \leftarrow$ 0.  \\ 
        
    }   
    \Else{
         \# \texttt{(3) the server broadcasts the parameters to clients.} \\ 
         Broadcast  (${W} (t)$, $b(t)$) to $M_t$ clients. \\ 
        \# \texttt{(4) Update each client's local parameters.} \\ 
         ${W}_{m}(t+1), b_{m}(t+1) \leftarrow$ update\_client\_parameters(${W}(t), b(t), T_c$) \\ 
        \# \texttt{(5) Update the server's parameters.} \\ 
        ${W}(t+1)$, ${b}(t+1)$ $\leftarrow$ aggregate\_parameters(${W_m}(t+1)$, ${b_m}(t+1)$, \textit{agg\_method}) \\ 
    }
}
\textbf{Return} (${W}(t+1), b(t+1))$
\end{algorithm}

\subsubsection{Multi-class Federated Logistic Regression (MFLR)}
To extend the BFLR to multi-class federated logistic regression (MFLR), we can adapt one-vs-rest (OVR for short, also known as one-vs-all (OVA)), which is a heuristic method utilizing binary classification algorithms for multi-class classification tasks \cite{galar2011overview}. 
More specifically, we assume there are $C$ target classes in the multi-class classification task we are facing.  During the learning phase, for each target class $c$, we fit a binary logistic regression model on the data with targets $Y_i \in \{1, 0\}$, where 1 represents the target class being $c$, while 0 represents the data being from the rest of the classes.  It's important to note that we can build all the $C$ BFLRs in parallel. The primary distinction between MFLR and BFLR is that in MFLR, each client will sent $C$ pairs of parameters $(W, b)$ to the server, and the server will broadcast all $C$ pairs of parameters $(W, b)$ to all the clients in each iteration.  
During the testing phase, for a given point $X \in \mathcal{R}^d$, we will run  all the classifiers and assign $X$ to the class $Y \in \mathcal{R}$ with the highest predicted probability, which can be seen below:
\begin{align}
     c^{*} & = \underset{c\ \in\ [C] }{\text{argmax}} \{P\{Y=c|X\}\} 
     = \underset{c\ \in\ [C]}{\text{argmax}} \{\sigma(W_{\{Y=c\}}^TX+b_{\{Y=c\}})\},
\end{align}
where $W_{\{Y=c\}}$ and $b_{\{Y=c\}}$ represent the coefficients and intercept of the model fitted on ``Class $c$ vs. Rest''.

\subsection{Logistic Regression Assumptions}
Logistic Regression relies on the following underlying assumptions \cite{harris2021primer, long2008crux, peng2002introduction, schreiber2018logistic}:
\begin{itemize}

    \item Linearity of the Logit: Continuous predictors are linearly related to a transformed version of the outcome (linearity).

    \item No or minimal multicollinearity: There should not be perfect multicollinearity among the independent variables.
    


    \item A large sample size for each class:
    A general guideline in \cite{vittinghoff2007relaxing} suggests to have  at least 10 observations with the least frequent outcome for each independent variable in the model. 
    
\end{itemize}

\section{State-of-the-Art Algorithms and Evaluation Metrics}
\label{sec:evaluation_metrics}

\subsection{Classical Algorithms}
Taking into account both efficiency and interpretability of the model, we present three supervised machine learning algorithms used in our experiments for numerical comparison, which are Logistic Regression (LR), Decision Tree (DT), and K-Nearest Neighbors (KNN).

\subsection{Evaluation Metrics}
In this section, we introduce commonly used metrics to 
evaluate performance of different supervised machine learning algorithms on both binary and multi-class classification tasks.

\begin{itemize}
    \setlength\itemsep{0.5em}

    \item \textbf{Accuracy}: The most intuitive metric to evaluate the model performance under the supervised machine learning setting (due to the existence of ground truth), which can be represented as below: 
    $$ACC = \frac{1}{N} \sum_{i=1}^{N} \mathbf{1}\{{\hat{Y}} = Y\},$$ where $\hat{Y}$ is the predicted class generated by the model, and $Y$ is the true class. For binary and multi-class classification tasks, we use a unified definition of accuracy.

    \item \textbf{$F_1$}:  The $F_1$ score is the harmonic mean of the precision and recall \cite{fscore2023}. It thus symmetrically represents both precision and recall in one metric. The more generic F score applies additional weights, valuing one of precision or recall more than the other.
    The highest possible value of an F-score is 1.0, indicating perfect precision and recall, and the lowest possible value is 0, if either precision or recall are zero.
    $${\displaystyle F_{1}={\frac {2}{\mathrm {Recall} ^{-1}+\mathrm {Precision} ^{-1}}}=2{\frac {\mathrm {Precision} \cdot \mathrm {Recall} }{\mathrm {Precision} +\mathrm {Recall} }}},$$
     where Precision and Recall are then defined as:
    $${\displaystyle {\begin{aligned}{\text{Precision}} ={\frac {\mathrm{TP}}{\mathrm{TP}+\mathrm{FP}}},\ {\text{Recall}} ={\frac {\mathrm{TP}}{\mathrm {TP}+\mathrm {FN}}}\,\end{aligned}}},$$ where $\mathrm{TP}$ is the true positive rate, $\mathrm{FP}$ is the false positive rate, and $FN$ is the false negative rate.
    
    In general, we use $F_1$ for binary classification tasks. However, we can also extend it to multi-class classification tasks. 
    In a multi-class classification task, for each class $c$ (treating this specific class as the positive class (labeled as 1) while labeling all other classes as the negative class (labeled as 0)), we compute the $F_{1c}$ on a binary task, and then average all the $F_{1c}$s to obtain the final $F_1$ as follows:  
    $$F_{1macro} = \frac{1}{C} \sum_{c=1}^{C} F_{1c},$$
    where $C$ is the number of classes, and $F_{1c}$ represents the $F_{1}$ of the $c$th class (i.e., $Y=1$ when $Y=c$ otherwise 0, which is the vanilla $F_{1}$ of a binary classification)
    Hereafter, unless otherwise stated, we use $F_{1}$ to represent both binary and multi classification ($F_{1macro}$) metric for simplicity. 

    \item \textbf{AUC}: The area under the receiver operating characteristic curve (ROC), i.e., AUC \cite{auc2023}. 
    The ROC curve is the plot of the true positive rate (TPR) against the false positive rate (FPR), at various threshold settings, and 
    AUC provides an aggregate measure of performance across all possible classification thresholds.
    It's a better evaluation metric for evaluating a classifier's performance than Accuracy (especially for the case where the data is imbalanced). Thus, we choose it as our main performance evaluation metric for comparison across different algorithms. As in $F_1$, we can also extend AUC to multi-class classification tasks. In a multi-class classification task, for each class $c$, we compute the $\mathrm{AUC_c}$ on a binary task, and then average all the $\mathrm{AUC_c}$s to obtain the final AUC as follows:  
    $$\mathrm{AUC_{marco}} = \frac{1}{C} \sum_{c=1}^{C} \mathrm{AUC_c},$$
    where $C$ is the number of classes, and $\mathrm{AUC_c}$ represents the AUC of the $c$-th class (i.e., $Y=1$ when $Y=c$ otherwise 0, which is the vanilla AUC of a binary classification)
    Unless otherwise stated, we use AUC to represent both binary and multi-class classification ($\mathrm{AUC_{macro}}$) metric for simplicity. 
    
\end{itemize}

\section{Experiments}
\label{sec:experiments}
In this section, we evaluate our proposed framework across multiple publicly available datasets, and compare its performance with classical algorithms as mentioned in Section \ref{sec:evaluation_metrics}. 
We make all datasets and corresponding source code publicly available at \href{https://github.com/kun0906/flr}{FLR}.

\subsection{Datasets}
First, we introduce four public financial datasets with various number of features and classes used in subsequent experiments. 
These datasets comprise two binary classification tasks and two multi-class classification tasks, as shown in Table \ref{tab:datasets}.

\subsection{Setup}
\vspace{-10pt}

\noindent\paragraph{Data Generation In Federated Learning Framework}
Here, we consider two data generation scenarios for the federated learning framework with and without outliers.

\textbf{Without Outliers}: In this case, we assume that there is no any adversary presented throughout the entire learning phase, meaning there are no outliers in the dataset.

\textit{Independent and Identically Distributed (IID)}: 
First, we shuffle the dataset and then partition the training set into $M$ subsets in a stratified manner based on the class ``Y''. 
Each client is then assigned one subset randomly, with each containing $n$ data points. Thus, the total number of data points is $n*M$.
We refer to this data generation process for clients as IID because it ensures that each client's data follows the same distribution as the original data.  It is more of an ideal case than what we often encounter in practice.

\textit{non-IID}: In this case, we first sampled 100 points from all the classes for each client. Then we partition clients to $C$ groups, with each group consisting of $M_c = \lfloor M/C \rfloor$ clients. We assign each class (excluding $100 * M$ data points) to one of these groups, which results in each client within that group having an additional $\lfloor (N_c-100*M)/M_c \rfloor$ data points. Therefore, each client ends up having a total of $\lfloor (N_c-100*M)/M_c \rfloor + 100 * C$ points. 
As we can see, each client's data does not follow the same distribution as the original data, and we refer to this data generation process as non-IID.
\noindent\begin{table*}[!ht]
\scriptsize
\aboverulesep=0ex
\belowrulesep=0ex
\centering
\captionof{table}{Four public datasets with various number of features and classes for binary and multi-class classification tasks.} \label{tab:datasets}
\begin{tabular}{|m{1.7cm}|m{1.2cm}|m{5.8cm}|m{0.99cm}|m{0.9cm}|m{2.6cm}|} \toprule
Data Name & Data Owner & Description (Purpose)       & Features (Used)*          & No. of Rows   & Labels (Distribution)      \\ \midrule
\hline 
BankMarketing \cite{bankmarketing} \newline (Binary)       & Portuguese banking institution &  The data is related to direct marketing campaigns of a Portuguese banking institution. The marketing campaigns were based on phone calls.  \newline 
The goal is to predict whether a client will subscribe a term deposit or not. \newline
Our experiments only use ``bank-full.csv'' as input for all the algorithms, and the target variable is denoted as ``Y''.
& 10 (7)     & 45,211   & 2 Classes including \newline
            Yes (5,289) and \newline No (39,922)  \\ \hline
LoanPred \cite{loanprediction}  \newline (Binary)   & Univ.AI &  The dataset belongs to a Hackathon organized by ``Univ.AI'' that collected historic customer behavior. \newline  
The goal is to predict who possible defaulters are for the consumer loans product. \newline
Our experiments only use ``Training Data.csv'' as input for all the algorithms, and the target variable is denoted as ``Risk\_Flag''. 
& 13 (5)     & 252,000    & 2 Classes including \newline 0 (221,004) and \newline 1 (30,996) \\ \hline
%
\hline 
CreditScore \cite{creditscore}  \newline (Multi-class)   & Unknown global financial company& The data is collected by a global financial company, which includes basic details of clients and their credit-related information. \newline 
The goal is to build an intelligent framework to segregate the people into credit score classes.  \newline 
Our experiments only use ``train.csv'' as input for all the algorithms, and the target variable is denoted as ``Credit Score''.
& 28 (15)   & 100,000     & 3 Classes including \newline 
Poor (28,998), \newline Standard(53,174), and  \newline Good (17,828) \\ \hline
CreditRisk \cite{creditrisk} \newline  (Multi-class) & Unknown financial  company &  The dataset contains credit histories of the customers from a financial institution. \newline
The goal is to predict possible credit defaulters in advance and help the financial institution to take steps accordingly. \newline 
Our experiment only use ``loan.csv'' as input for all the algorithms, and the target variable is denoted as ``loan\_status''.
Note that the data has 10 classes in total; yet our focus is solely on the top six classes due to the scarcity of records in the remaining classes
& 74 (30)     & 887,379    & Top 6 Classes including  Current (601,779), Fully Paid (207,723) 
Charged Off (45,248), 
Late (31-120 days) (11,591), Issued (8,460), and In Grace Period (6,253). 
\\ \hline
\end{tabular}
\begin{flushleft}
 *  The number in the ``Features(Used)'' column represents the total number of columns and the number of features used in our experiments. 
\end{flushleft}
\end{table*}

\textbf{With Outliers}: 
In this section, we assume the presence of an adversary or adversaries during the entire learning phase. The adversary sends wrong or fake local parameters to the server in order to fail the global model. If the server uses the mean aggregation of the collected parameters, the global model won't function properly. Therefore, we apply different robust aggregation methods to address and rescue the model from the effects of outliers. 

\textit{IID}: 
First, we shuffle the dataset and then partition the training set into $M-\lfloor M * p_{out} \rfloor$ subsets in a stratified manner based on the class ``Y'', where $p_{out}$ is the percentage of adversaries.  
Each normal client is then assigned one of these subsets randomly, with each containing $n$ data points. Thus, the total number of data points is $n*(M-\lfloor M * p_{out} \rfloor)$. As before, we refer to this data generation process for clients as IID because each normal client's data follows the same distribution as the original data. The difference with the case of the IID data without outliers is that here we have $p_{out}\%$ of the partitions generated by adversaries. 

\textit{non-IID}: In this case, each client has $s$ points from a class and an additional 10\% of points from the remaining classes, i.e., each client has data sampled from all the classes.
We mainly emphasize evaluating our framework under this scenario because it is more likely to occur in practice than the IID case.

Furthermore, in this case (with outliers), we further consider three experimental settings to assess the performance of our proposed framework, which are as follows:
\begin{enumerate}
    \item {Different Sample Sizes ($s$)}: First, we investigate how different sample sizes affect the performance of each method. In this context, $s$ represents the number of normal data points associated with each client.
    We pick the sample size $s$ from [25, 50, 100, 150, 200], while keeping the number of clients $M$ fixed at 100 and the outlier percentage at $p_{out}=0.1$. 
    
    \item {Different Percentages of Outliers ($p_{out}$)}: Next, we assess how the strength of attack power influences the performance of each method. The more outliers, the stronger the attacker power is. We vary the outlier percentage $p_{out}$ from 0 to 0.2 with a step size of 0.5, while keeping the number of clients $M$ fixed at 100 and the sample size $s$ fixed at 100.

    \item {Different Number of Clients ($M$)}: We also examine how the number of clients impacts the performance of each method. We pick the number of clients $M$ from [10, 50, 100, 150, 200], while keeping the sample size $s$ fixed at 100 and the outlier percentage $p_{out}$ at 0.1.
\end{enumerate}

We summarize the different data generation cases in Table \ref{tab:data_generation}. 

\begin{table*}[htbp!]
\scriptsize
\aboverulesep=0ex
\belowrulesep=0ex
\centering
\captionof{table}{Data generation in federated learning framework.} 
\label{tab:data_generation}
\begin{tabular}{|p{3.6cm}|p{3.6cm}|p{3.6cm}|p{3.6cm}|}
\toprule
\multicolumn{2}{|c|}{All data without outliers }                     & \multicolumn{2}{c|}{Sampled data with outliers}  \\ 
\midrule
\multicolumn{1}{|c|}{IID} & \multicolumn{1}{c|}{non-IID} & \multicolumn{1}{c|}{IID} & \multicolumn{1}{c|}{non-IID}      \\ 
\midrule
{1. We randomly partition the dataset comprising $N$ data points into $M=100$ subsets, each containing $N/M$ points. This partitioning is carried out in a stratified manner based on the labels $y$.\newline

2. Each subset is assigned to a client. This partitioning ensures that every client has the same distribution as the population.}       
& {1. For each client, we first randomly select 100 data points from each class. \newline

2. Next, we partition clients to $C$ groups, and each group has $M_c = M/C$ clients. We assign each class (excluding $100 M$ data points) to a group, and each client in the group has another $(N_c-100M)/M_c$. Finally, each client has $(N_c-100M)/M_c + 100 C$ in total.
}
&{1. We randomly partition the dataset comprising $N$ data points into $M$ subsets, and this partitioning is carried out in a stratified manner based on the labels $y$. \newline

2. Next, for each client, we sample $s$ data point from each subset, and ensure that the sampled $s$ points has the same distribution as the population. 
}
&{1. For each client, we first randomly select $\lfloor s \cdot 0.1 \rfloor$ data points from each class. \newline

2. Next, we evenly partition each class into $M_c = M/C$ clients, ensuring that each client possesses $ s + \lfloor 0.1 \cdot s \cdot (C-1) \rfloor$ data points from all the classes. However, each client data does not follow   the same distribution as the population. 
}
\\ 
\midrule
\multicolumn{2}{|p{7cm}|}{Note that $M=100$, and we use all the data.} & 
\multicolumn{2}{p{7cm}|}{Note that we sample a subset of the data according to different $s$, $p_{out}$, and $M$. 
} \\
\bottomrule
\end{tabular}
\end{table*}
\noindent\paragraph{Algorithm Setting}
All the algorithms have been implemented using Python 3.9.7, along with Scikit-learn 1.0.2, and Numpy 1.22.1. Further details regarding the software environment are available in the ``requirements.txt'' of the anonymized repository.
All the algorithms use the default parameters in scikit-learn unless otherwise stated. 
For federated logistic regressions,  we set the maximum number of iterations (for communication between the server and the clients) to 100, and each client updates its local parameters to 10 iterations per round.

\noindent\paragraph{Result Report Setting}
We conduct each experiment 10 times with a range of random seeds spanning from 0 to 1000, incremented by 100.  We then calculated the average and standard deviation of the results for ACC, F1, and AUC, as mentioned in Section \ref{sec:evaluation_metrics}.
Finally we report the results in the format of $\mu \pm 1.96 \sigma$. 

\subsection{Results}

\paragraph{Assumption Verification}
First, we verify the assumptions for LR on the datasets.  

\textbf{Data Size Verification}:
Although there are no precise guidelines for the number of observations/data points required,
a general rule in \cite{vittinghoff2007relaxing} suggests a minimum ratio of 10 observations per predictor variable, with a minimum sample size of 100. 
The reasoning behind this guideline is to reduce the risk of overfitting, where the model may fit the noise in the data rather than the underlying relationships. With more data points, the model is less likely to overfit and more likely to generalize well to new, unseen data, i.e., we can achieve more stable and reliable results. 
For example, if you have $m$ predictor variables, it is generally recommended to have at least $10 m$ observations. 
From Table \ref{tab:datasets}, we can easily verify that all our datasets meet or exceed the general rule recommendations. Therefore, the results reported in these experiments are considered stable.

\begin{figure*}[!htbp]
  \centering \captionsetup{hypcap=false} 
\includegraphics[height=13.cm,keepaspectratio]{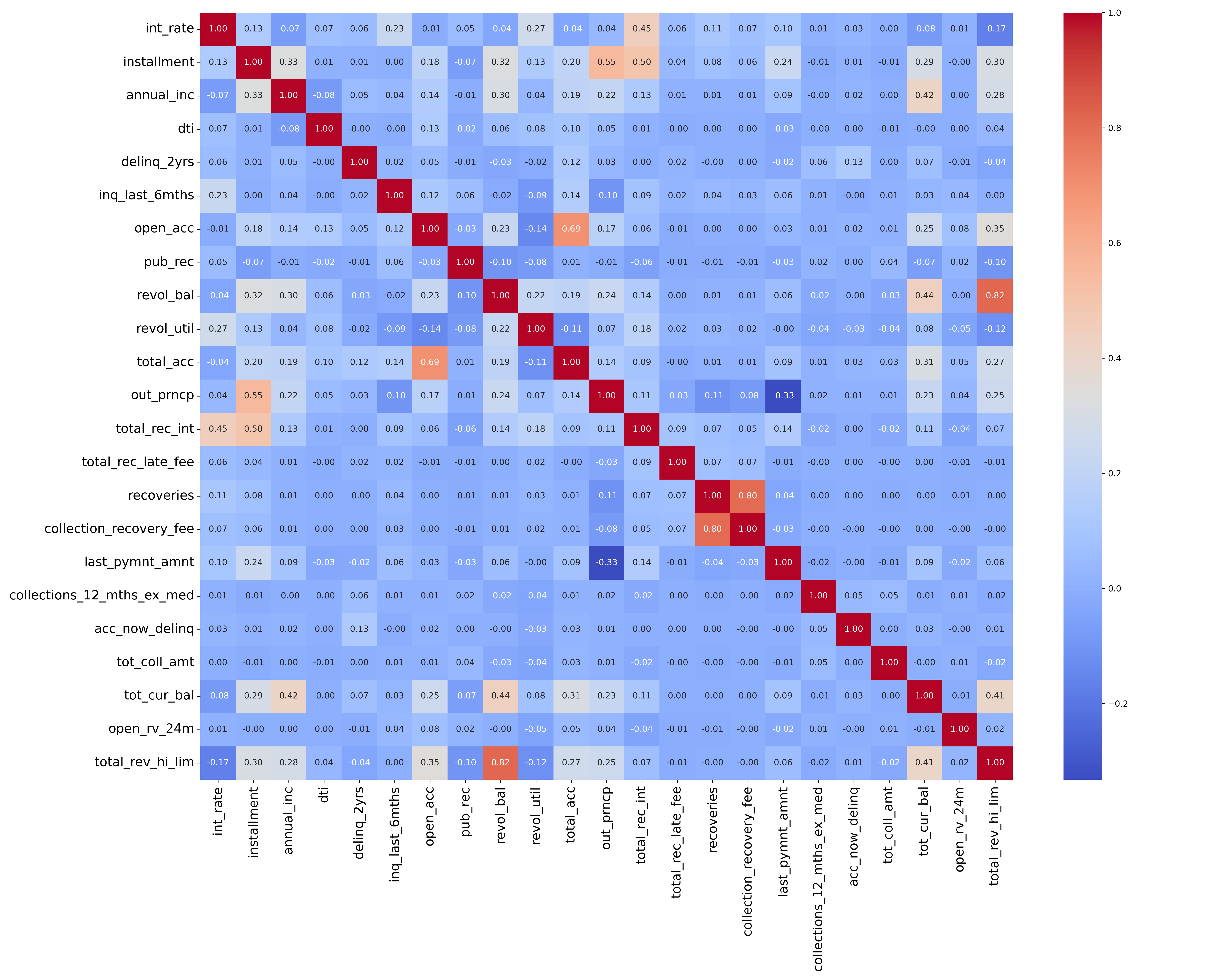}
  \caption{Feature correlation of CreditRisk after iteratively removing the features with higher VIFs (>10).}
  \label{fig:correlation_vif}
\end{figure*}

\textbf{Multicollinearity Verification}:
We use variance inflation factor (VIF) \cite{o2007caution} to assess multicollinearity in features. VIF quantifies how much the variance of the estimated regression coefficients is increased due to multicollinearity. High VIF values indicate that the predictor variables are highly correlated, which can make it difficult to interpret the individual effects of each variable.
%
A common threshold is VIF $> 10$ \cite{o2007caution}, but the specific threshold can vary based on the context.
We iteratively assess and remove predictor variables with high VIF values until we achieve a model with reasonable multicollinearity, e.g., all VIFs are less than 10. After removing the features that have high VIF values, we visualize the correlation of features to further check if there still is correlation between each two features. 
For instance, 
Fig. \ref{fig:correlation_vif} shows the correlations of features of CreditRisk after iteratively removing the features with high VIFs ($>10$). The redder the color, the higher the correlation. From the results, we don't see strong correlation between each two features. Also, there is no multicollinearity in features (verified by VIF), therefore, we can use these features as input for logistic regression.  
Moreover, in the subsequent experiments, we use regularization (with $\ell_2$ penalty as a default parameter in scikit-learn for logistic regression) to further alleviate the effect of multicollinearity to logistic regression. 

\textbf{Linearity}:
LR assumes the logit of the response variable $Y$ is a linear function of the features $X$. To assess this for a binary classification task, we can employ the Box-Tidwell test \cite{fox2016}.  For each of our features $X_i$, we calculate the interaction between the feature and its natural logarithm, meaning we calculate $Z_i=X_i ln(X_i)$.  We then regress the response variable $Y_i$ on $X_i$ and $Z_i$.  Our null hypothesis is that the linearity assumption holds, meaning that the regression coefficient associated to $Z_i$ is zero.  If the corresponding $p$-value falls below our significance level of 0.01, then we reject the linearity assumption.  
Applying this to the BankMarketing dataset, we find that the linearity assumption fails for all features.
When looking at the LoanPred dataset, we fail to reject the linearity assumption for the CURRENT\_JOB\_YRS and CURRENT\_HOUSE\_YRS features, while we reject the linearity assumption for the other features.  

\paragraph{Baseline}
\label{sec:baseline}
After assumption verification, we first report the results achieved by LR, DT, and KNN in the centralized setting, where we assume that all the data without any outliers resides in one location, and we can fit and evaluate models on all the data and the same location without worrying about any costs (such as communication cost between server and client) of collecting all the data together. 
Table \ref{tab:baseline} shows the testing performance in terms of ACC, $F_1$, and AUC ($\mu \pm 1.96\sigma$) obtained by LR, DT, and KNN across all four datasets in both binary and multi-class classification tasks. 

From the results, we observe that all three methods have generally achieved superior AUC performance compared to ACC and $F_1$ score. This can be attributed to AUC's comprehensive assessment across various thresholds, whereas ACC and $F_1$ exclusively focus on the default thresholds.
Furthermore, a crucial factor contributing to this distinction is that ACC and $F_1$ tend to be highly affected by imbalanced data compared to AUC. More details of class distribution can be seen in Table \ref{tab:datasets}.

Furthermore, we notice that DT and KNN are not always better LR in terms of AUC.
The underlying reason for this phenomenon may be attributed to the presence of linear relationships between the predictors (or features) and the logit of the response variable. LR excels in capturing these linear relationships relative to DT and KNN.

\begin{table*}[htbp!]
\scriptsize
\aboverulesep=0ex
\belowrulesep=0ex
\centering
\captionof{table}{Baseline of ACC, $F_1$, and AUC ($\mu \pm 1.96 \sigma$) obtained by LR, DT, and KNN on different datasets under the centralized setting.} 
\label{tab:baseline}
\begin{tabular}{|l|c|l|l|l|l|} 
\toprule
\multirow{2}{*}{Metrics} & \multirow{2}{*}{Algorithms} & \multicolumn{2}{c|}{Binary}                     & \multicolumn{2}{c|}{Multi-class}  \\ 
\cmidrule{3-6}
                         &                             & \multicolumn{1}{l|}{BankMarketing} & \multicolumn{1}{l|}{LoanPred} & \multicolumn{1}{l|}{CreditScore} & \multicolumn{1}{l|}{CreditRisk}      \\ 
\midrule
\multirow{2}{*}{ACC}     & LR                          & 0.69 $\pm$ 0.01        & 0.49 $\pm$ 0.01                   &    0.50 $\pm$ 0.00                   &    0.88 $\pm$ 0.00      \\ 
                         & DT                          &   0.86 $\pm$ 0.01                  & 0.88 $\pm$ 0.00                   &                0.69 $\pm$ 0.01          &    0.94 $\pm$ 0.00       \\ 
                        & KNN                          &   0.88 $\pm$ 0.01                 & 0.89 $\pm$ 0.00                   &                0.59 $\pm$ 0.00          &    0.91 $\pm$ 0.00       \\ 
                         
\midrule
\multirow{2}{*}{F1}      & LR                          & 0.39 $\pm$ 0.01                   & 0.22 $\pm$ 0.00                    &     0.50 $\pm$ 0.00                   &     0.49 $\pm$ 0.00      \\ 
                         & DT                          & 0.37 $\pm$ 0.01                   & 0.62 $\pm$ 0.01                   &                 0.67 $\pm$ 0.01       &     0.58 $\pm$ 0.01      \\ 
                         & KNN                          &   0.38 $\pm$ 0.02                 & 0.53 $\pm$ 0.01                    &                0.53 $\pm$ 0.01          &    0.47 $\pm$ 0.00       \\ 
\midrule
\multirow{2}{*}{AUC}     & LR                          & 0.83 $\pm$ 0.02                   & 0.53 $\pm$ 0.01                   &        0.74 $\pm$ 0.00                &      0.89 $\pm$ 0.00     \\ 
                         & DT                          & 0.64 $\pm$ 0.02                   & 0.85 $\pm$ 0.00                   &                 0.75 $\pm$ 0.01       &     0.78 $\pm$ 0.00      \\
                         & KNN                          &   0.79 $\pm$ 0.02                 & 0.87 $\pm$ 0.01                   &                0.73 $\pm$ 0.00          &    0.79 $\pm$ 0.00       \\ 
\bottomrule
\end{tabular}
\end{table*}

\paragraph{Federated LR without Outliers}
As previously mentioned, the assumption of having access to all the data is not always accurate. Therefore, federated learning frameworks have been introduced as a solution. In this section, we present the performance of FLR with mean aggregation on the server and compare it to more robust strategies, such as median and trimmed mean mentioned in Section \ref{sec:proposed_method}. The results (as shown in Table \ref{tab:no_adversary}) demonstrate that without any outliers, all FLRs achieve nearly identical performance. Employing more advanced strategies does not lead to performance improvement because, in the absence of outliers, mean aggregation proves to be the optimal choice.

Furthermore, we have achieved similar performance in AUC on IID and non-IID data, but we get slightly better performance on IID than non-IID dta. In this case, we have 100 clients and there are no outliers (i.e., $p=0.0$). 
Moreover, we can see that our proposed framework can achieve comparable performance in terms of AUC with the baseline performance obtained by LR, DT and KNN in Table  \ref{tab:baseline} when there are no outliers.

\noindent\begin{table*}[!htbp]
\scriptsize
\aboverulesep=0ex
\belowrulesep=0ex
\centering
\captionof{table}{ACC, F1, and AUC ($\mu \pm 1.96 \sigma$) 
 obtained by FLR on different datasets under the federated setting with IID and non-IID data, with 100 clients, but no outliers (i.e., $p=0.0$).} \label{tab:no_adversary}
\begin{tabular}{|l|l|l|c|l|l|l|} 
\toprule
\multirow{2}{*}{Data Distribution} & \multirow{2}{*}{Metrics} & \multirow{2}{*}{Algorithms} & \multicolumn{2}{c|}{Binary}                     & \multicolumn{2}{c|}{Multi-class}  \\ 
\cmidrule{4-7}
                         &   &                           & \multicolumn{1}{l|}{BankMarketing} & \multicolumn{1}{l|}{LoanPred} & \multicolumn{1}{l|}{CreditScore} & \multicolumn{1}{l|}{CreditRisk}      \\ 
\midrule
\multirow{9}{*}{IID} & \multirow{3}{*}{ACC} & FLR-mean       & 0.69 $\pm$ 0.01 & 0.49 $\pm$ 0.01 & 0.50 $\pm$ 0.00 & 0.88 $\pm$ 0.00\\   
    &  & FLR-median       & 0.69 $\pm$ 0.01 & 0.49 $\pm$ 0.01   &  0.50 $\pm$ 0.00 & 0.88 $\pm$ 0.00\\
    &  & FLR-trim\_mean       & 0.69 $\pm$ 0.01 & 0.49 $\pm$ 0.01 &  0.50 $\pm$ 0.00 & 0.88 $\pm$ 0.00\\ 
\cmidrule{2-7}
& \multirow{3}{*}{F1} & FLR-mean       & 0.39 $\pm$ 0.01 & 0.22 $\pm$ 0.00 &    0.50 $\pm$ 0.00  & 0.49 $\pm$ 0.00  \\
    &  & FLR-median       & 0.39 $\pm$ 0.01 & 0.21 $\pm$ 0.00 &  0.50 $\pm$ 0.00    &  0.49 $\pm$ 0.00 \\
    &  & FLR-trim\_mean       & 0.39 $\pm$ 0.01 & 0.22 $\pm$ 0.00 &  0.50 $\pm$ 0.00   & 0.49 $\pm$ 0.00  \\ 
\cmidrule{2-7}
& \multirow{3}{*}{AUC} & FLR-mean       & 0.83 $\pm$ 0.02 & 0.53 $\pm$ 0.01 & 0.74 $\pm$ 0.00 & 0.89 $\pm$ 0.00 \\ 
    &  & FLR-median       & 0.83 $\pm$ 0.01 & 0.53 $\pm$ 0.01 & 0.74 $\pm$ 0.00  &   0.89 $\pm$ 0.00  \\
    &  & FLR-trim\_mean       & 0.83 $\pm$ 0.02 & 0.53 $\pm$ 0.01 &   0.74 $\pm$ 0.00  &  0.89 $\pm$ 0.00 \\
\hline
\hline
\multirow{9}{*}{NON-IID} & \multirow{3}{*}{ACC} & FLR-mean       & 0.70 $\pm$ 0.01 & 0.49 $\pm$ 0.04 & 0.48 $\pm$ 0.02 & 0.55 $\pm$ 0.13\\       
    &  & FLR-median       & 0.70 $\pm$ 0.01 & 0.50 $\pm$ 0.03   & 0.47 $\pm$ 0.02 & 0.34 $\pm$ 0.03\\ 
    &  & FLR-trim\_mean       & 0.70 $\pm$ 0.01 & 0.49 $\pm$ 0.04 & 0.48 $\pm$ 0.02 & 0.56 $\pm$ 0.12\\ 
\cmidrule{2-7}
& \multirow{3}{*}{F1} & FLR-mean       & 0.39 $\pm$ 0.01 & 0.21 $\pm$ 0.00 &   0.48 $\pm$ 0.02  & 0.39 $\pm$ 0.04 \\
    &  & FLR-median       & 0.39 $\pm$ 0.01 & 0.21 $\pm$ 0.00 &   0.48 $\pm$ 0.02  &  0.32 $\pm$ 0.01 \\
    &  & FLR-trim\_mean       & 0.39 $\pm$ 0.01 & 0.21 $\pm$ 0.00 &    0.48 $\pm$ 0.02 &  0.39 $\pm$ 0.04 \\ 
\cmidrule{2-7}
& \multirow{3}{*}{AUC} & FLR-mean       & 0.83 $\pm$ 0.02 & 0.53 $\pm$ 0.01 & 0.73 $\pm$ 0.01 & 0.87 $\pm$ 0.00\\ 
    &  & FLR-median       & 0.83 $\pm$ 0.01 & 0.53 $\pm$ 0.01 & 0.73 $\pm$ 0.01 &   0.88 $\pm$ 0.01  \\
    &  & FLR-trim\_mean       & 0.83 $\pm$ 0.02 & 0.53 $\pm$ 0.01 &  0.73 $\pm$ 0.01 &  0.87 $\pm$ 0.00 \\ \hline

\end{tabular}
\end{table*}

\paragraph{Federated LR with Outliers} 
In this section, we further evaluate the models performance in the presence of outliers, where we sample subsets of the data for different experimental settings, and more details can be seen in Table \ref{tab:data_generation}. 

\begin{figure}[htbp!]
    \centering
    \begin{minipage}{0.32\textwidth}
        \begin{center}
            \includegraphics[scale=0.33]{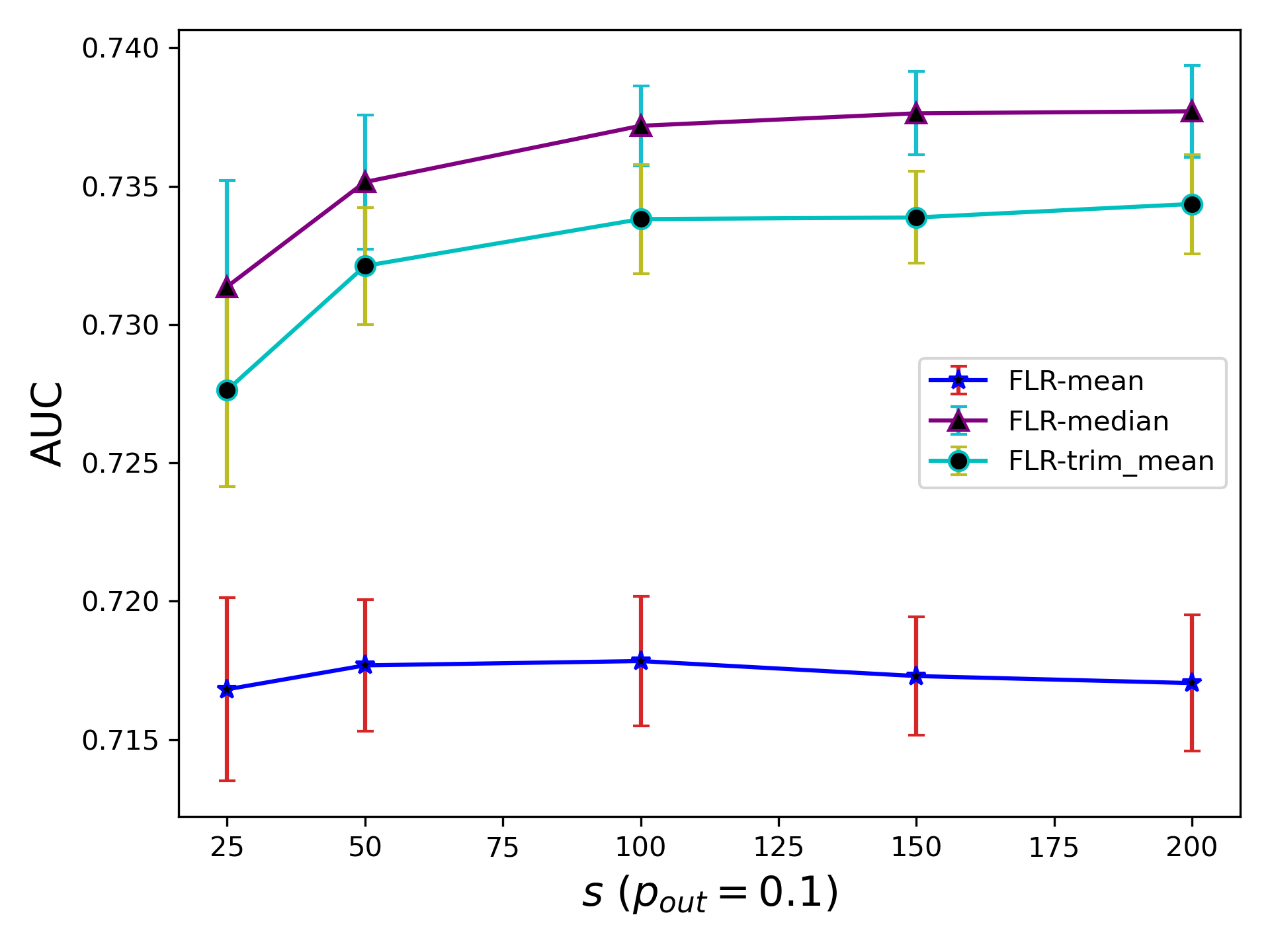}
        \subcaption{The effect of different sampling sizes $s$ with $M=100$ clients and $p_{out}=0.1$ on non-IID data.}
         \label{fig:creditscore_s_non_iid}
        \end{center}
    \end{minipage}\hfill
    \begin{minipage}{0.32\textwidth}
        \begin{center}
            \includegraphics[scale=0.33]{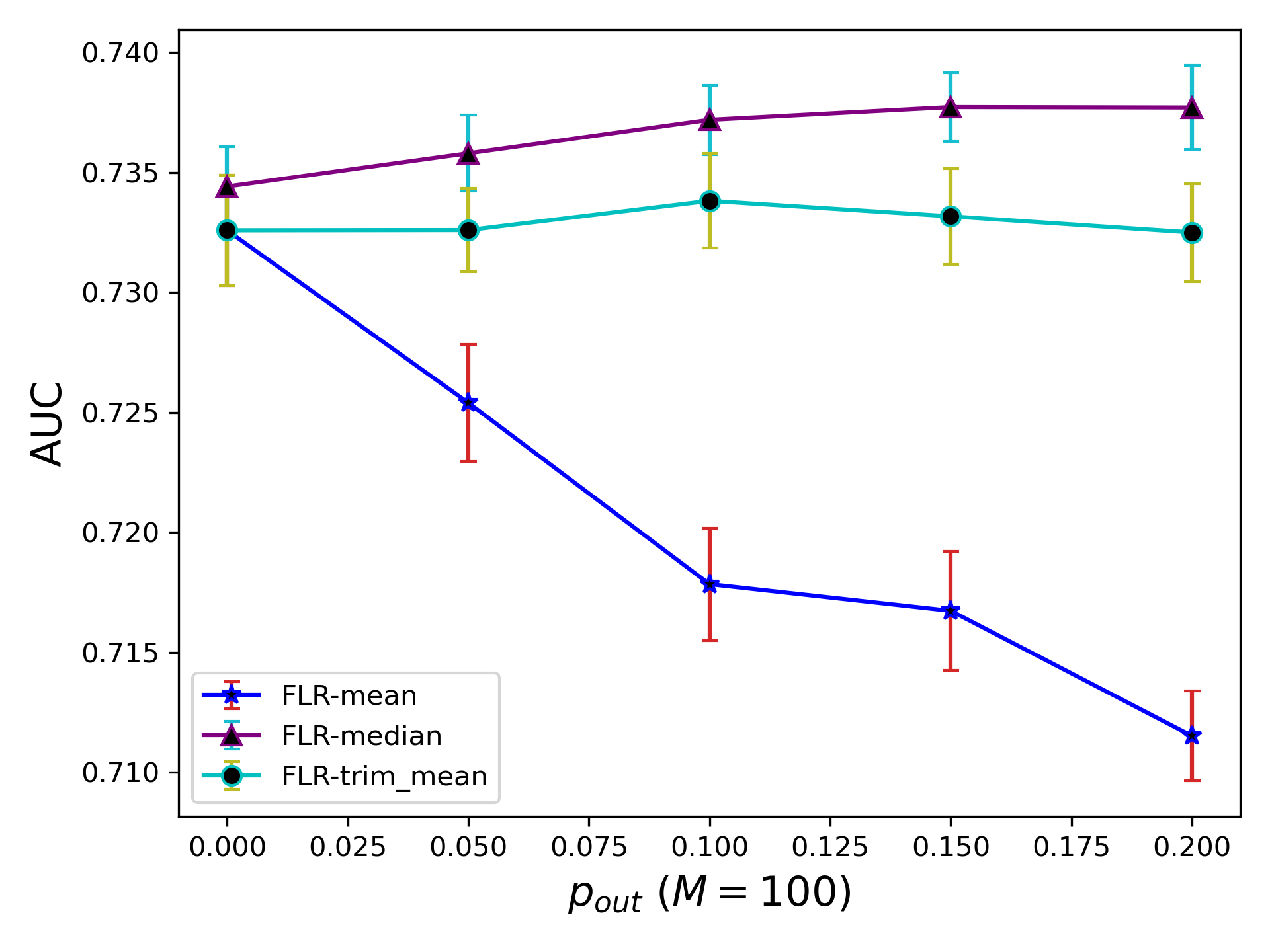}
       \subcaption{The effect of number of clients $M$ with fixed $p_{out}=0.1$ and $s=100$ on non-IID data.}
       \label{fig:creditscore_p_non_iid}
        \end{center}
    \end{minipage} \hfill
    \begin{minipage}{0.32\textwidth}
        \begin{center}
            \includegraphics[scale=0.33]{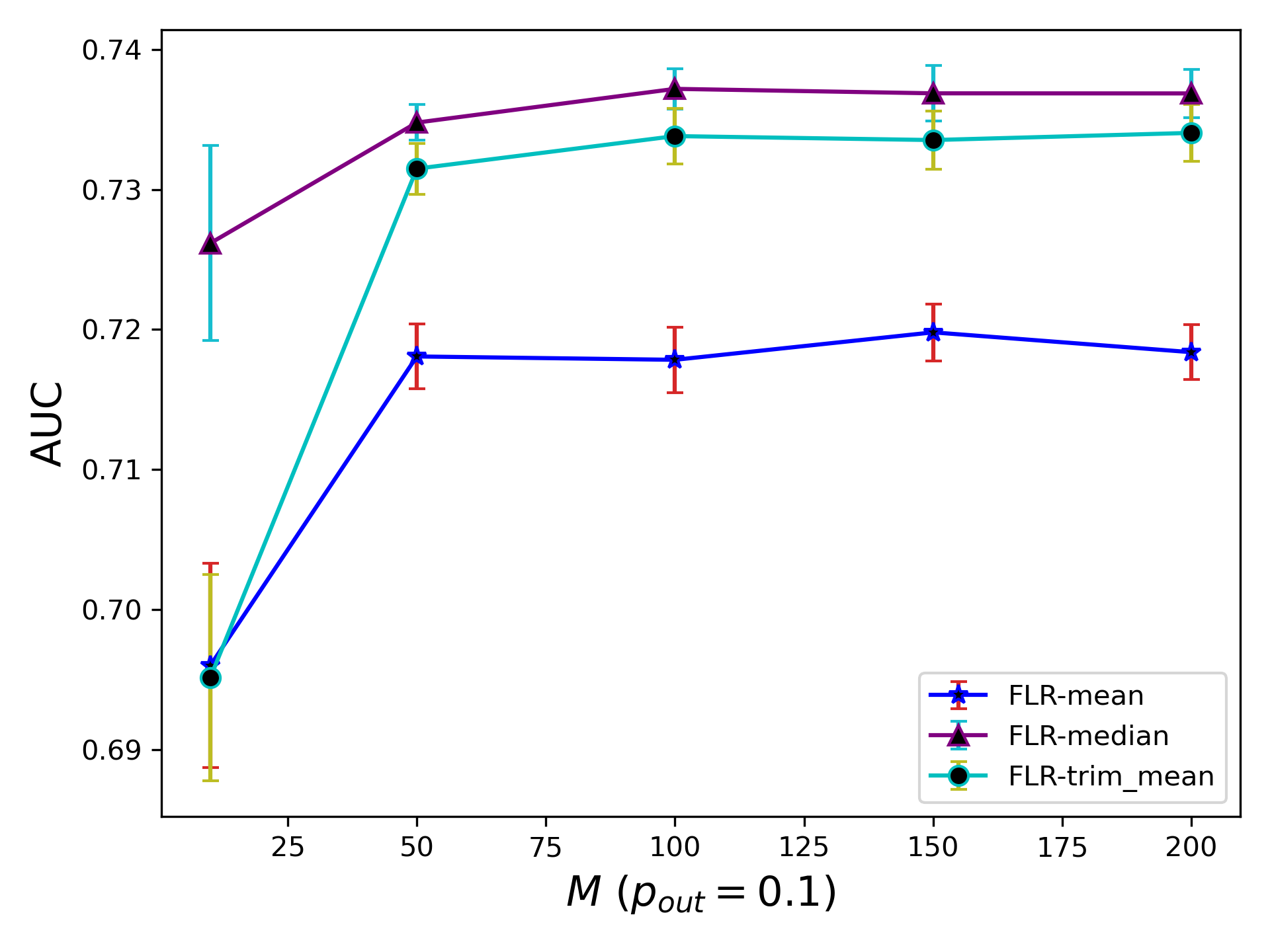}
       \subcaption{The effect of number of clients $M$ with fixed $p_{out}=0.1$ and $s=100$ on non-IID data.}
       \label{fig:creditscore_C_non_iid}
        \end{center}
    \end{minipage}
    \caption{Different cases on non-IID data.}
\end{figure}

First, we can easily see in Fig. \ref{fig:creditscore_s_non_iid} that as the sample size $s$ increases (but $M$ and $p_{out}$ are fixed, where $M=100$ and $p_{out}=0.1$), the performance of FLR with median and trimmed mean aggregation also increases, although not by much, when $s$ is less than 100. When $s$ exceeds 100, their performance tends to be stable on non-IID data, and this is also true for IID data.

Furthermore, FLR with mean aggregation has the worst performance in AUC than any robust aggregation FLRs due to the presence of outliers. 
And as we increase the outlier percentage, this phenomenon becomes even more pronounced for FLR with mean aggregation, as illustrated in Fig. \ref{fig:creditscore_p_non_iid}. 
However, FLRs with median and trimmed mean aggregation exhibit considerably greater enhancements compared to FLR with mean aggregation. The reason behind this is due to the high susceptibility of mean aggregation to outliers. Even a single outlier can profoundly distort the mean in cases where the outlier lies far from the normal data. Consequently, FLR with mean aggregation becomes unreliable. However, FLRs with median and trimmed mean aggregation display greater robustness to outliers, resulting in minimal performance degradation as the outlier percentage increases.

Moreover, we notice that FLR with median aggregation has a better performance than FLR with trimmed-mean aggregation across all the datasets. The reason is that after filtering $p_{out}\%$ (same as the percentage of outliers) of data points, there might still are a few large inliers can drive the trimmed mean far from the true mean; however, FLR with median aggregation is less effect by the outliers. 
Also, we also notice that the proposed framework has a slightly better performance on the IID data than non-IID data, which is reasonable because the client can learn more stable parameters on IID data than that of non-IID data. 
Additionally, we've observed that the number of clients doesn't significantly impact the performance of each FLR especially when $M$ is greater than or equal to 50, which can be seen in Fig. \ref{fig:creditscore_C_non_iid}.

\begin{figure}[!htbp]
    \centering
    \begin{minipage}{0.32\textwidth}
        \begin{center}
            \includegraphics[scale=0.33]{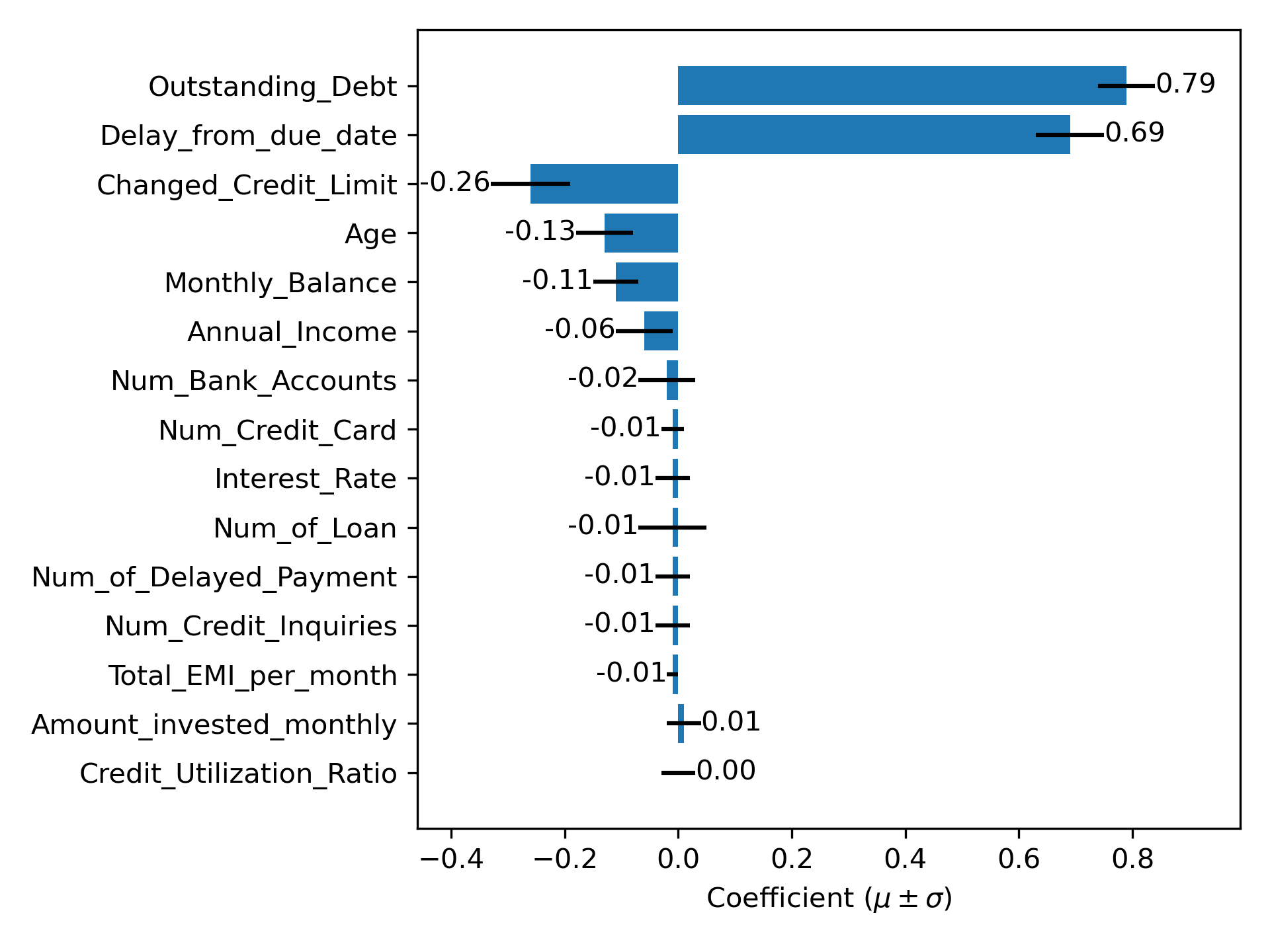}
        \subcaption{Poor vs. Rest}
        \label{fig:poor_vs_rest}
        \end{center}
    \end{minipage}\hfill
    \begin{minipage}{0.33\textwidth}
        \begin{center}
            \includegraphics[scale=0.33]{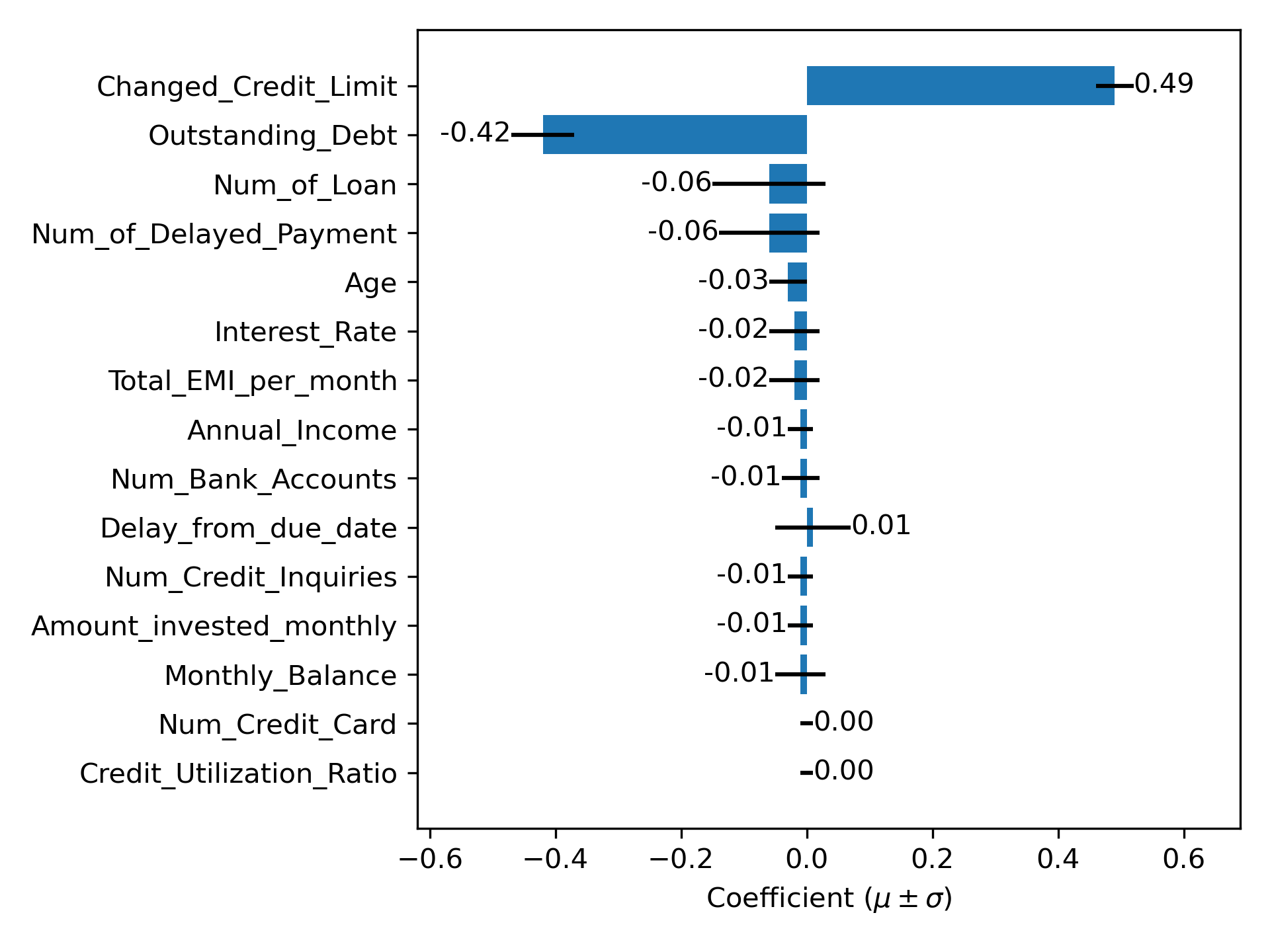}
       \subcaption{Standard vs. Rest}
        \label{fig:standard_vs_rest}
        \end{center}
    \end{minipage} \hfill
    \begin{minipage}{0.33\textwidth}
        \begin{center}
            \includegraphics[scale=0.33]{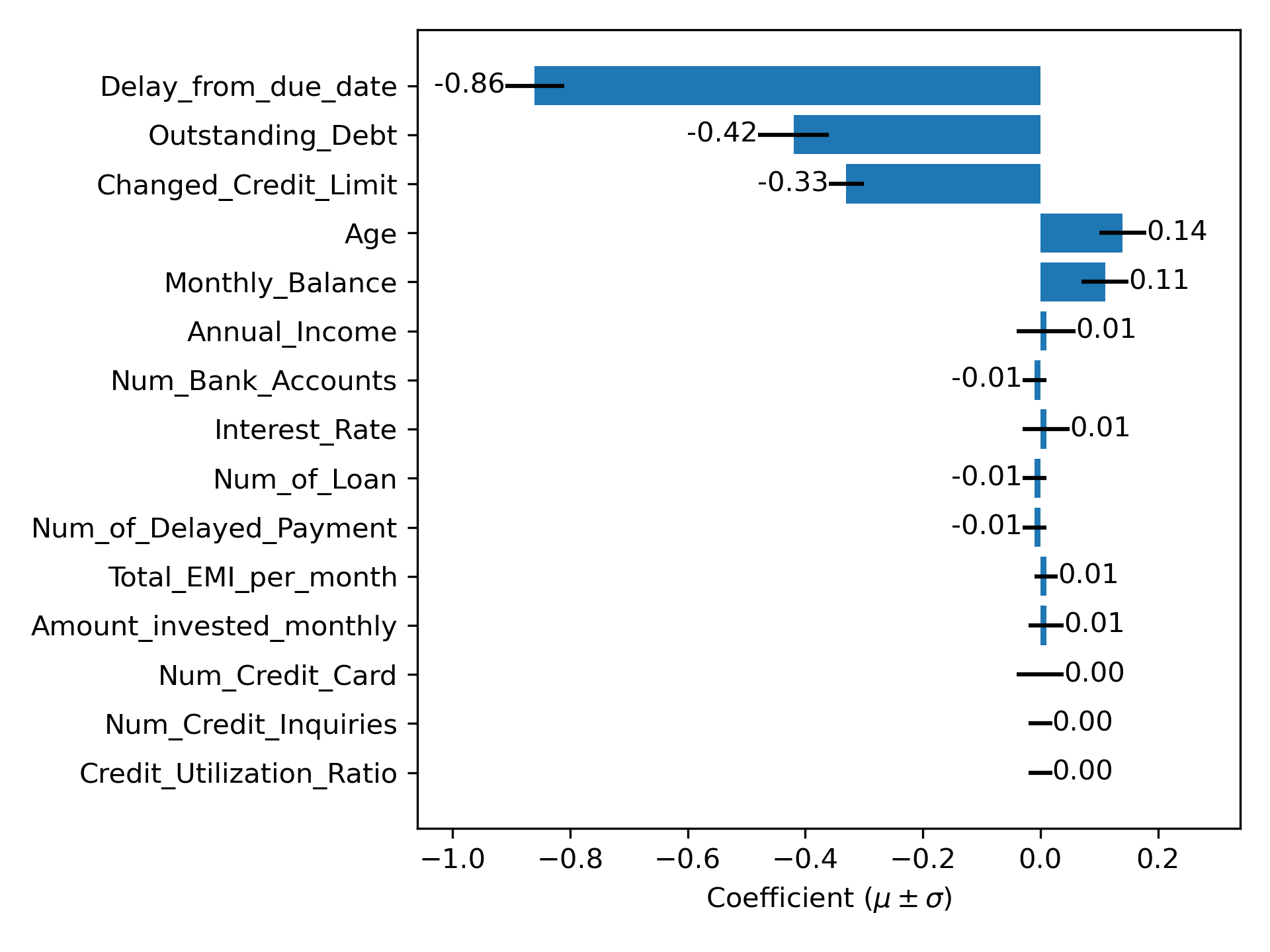}
       \subcaption{Good vs. Rest}
        \label{fig:good_vs_rest}
        \end{center}
    \end{minipage}
    \caption{Feature importances obtained by FLR with median aggregation on CreditScore with non-IID data, but no outliers.}
    \label{fig:feature_importance_non_iid_no_outliers}
\end{figure}

\paragraph{Feature Importance Report}
Here, we analyze the feature importance obtained by our proposed framework without outliers.  More specifically, we focus on analyzing the coefficients of each feature obtained by our proposed FLR framework in a multi-class classification task. According to Table \ref{tab:no_adversary}, it becomes evident that our proposed FLR framework, with median and trimmed\_mean aggregation, demonstrates nearly the same performance as the vanilla FLR with mean aggregation across all the evaluation metrics on both binary and multi-class classification tasks. Therefore, we mainly focus on the results obtained by the FLR with median aggregation. Additionally, due to space constraints, we only report the results on CreditScore with non-IID data, but no outliers. For IID data, we observe the similar results.
%
In the context of a multi-class classification task, we employ an one-vs-rest approach, resulting in individual binary logistic regression models for each class. For each of these models, we present the feature importances based on their corresponding coefficients, which can be seen in Fig. \ref{fig:feature_importance_non_iid_no_outliers}. Here, for each feature, we've computed both the mean and standard deviation of the corresponding coefficient across 10 repeated experiments. And on this figure, the X-axis represents the coefficients of each feature in $\mu \pm 1.96 \sigma$, while the Y-axis corresponds to the corresponding features. Additionally, the coefficients have been arranged in descending order for each class, disregarding whether they are positive or negative values.
The results in Fig. \ref{fig:feature_importance_non_iid_no_outliers} reveal that different models assign varying degrees of importance to different features. For example, ``Delay\_from\_due\_date'' is the most important feature (with a coefficient of -0.86) for the logistic regression model fitted on ``Good vs. Rest'', as shown in Fig. \ref{fig:good_vs_rest}, which is also significant (with a coefficient of 0.69) for the logistic model with ``Poor vs. Rest'' in Fig. \ref{fig:poor_vs_rest}. However, for the logistic regression model fitted on ``Standard vs. Rest'' in Fig. \ref{fig:standard_vs_rest}, its effect becomes very insignificant (with a coefficient of 0.01).
On the other side, certain features exhibit consistent importance across all three models. For instance, ``Outstanding\_Debt'' and ``Changed\_Credit\_limit'' consistently rank among the top three features in all three logistic regression models, regardless of whether we're considering IID or non-IID data, although their coefficient values vary among the models.
Moreover, we also observe that certain features appear to have very limited impact to all the models, such as ``credit\_utilization\_ratio'', which has nearly negligible impact on all three models.

\vspace{-0.4cm}
From the above analysis of the coefficients obtained by FLR, we can easily explain which features and why they play a crucial role in the performance of each model.
We would like to emphasize that the characteristics of FLR are very helpful and important for making decisions compared to complex models, such as random forests and neural networks.

\section{Discussion}
\label{sec:discussion}
In this paper, our primary focus
centers on the validation and assessment of the feasibility of our proposed framework and the interpretability of our model.
Consequently, there are certain aspects that fall outside the scope of this study, and we discuss them in this section. 

\textbf{Sophisticated/tailored Robust Federated Framework}: We would like to emphasize that our work introduces the application of the federated learning framework to tackle financial challenges, such as loan prediction. 
Our focus is to address more general issues under the federated framework (considering data distributed across the edge clients), rather than providing a sophisticated and tailored framework (with rigorous theoretical analysis) to a special question under very constricted assumptions \cite{zhu2023byzantine, pillutla2022robust}. For example, using Tukey median to aggregate the parameters on the server in federate learning \cite{tukey1977exploratory}. 
Additionally, there are other sophisticated attack strategies to generate outliers \cite{jere2020taxonomy}.
These aspects should be further investigated to the performance of our framework, and how to optimize the proposed framework to address the corresponding scenarios would be more important.

\textbf{Hyperparameter Tuning}: This work does not delve into the impact of hyperparameter tuning on the model's performance \cite{yang2020hyperparameter}. 
However, we acknowledge that fine-tuning hyperparameters can enhance model performance, and we defer this aspect to future research. For instance, we will consider issues such as how to configure the learning rate, determine the number of iterations, and choose appropriate regularization techniques for logistic regression. Furthermore, in this work, we randomly initialize the proposed framework without leveraging any local data information. Some existing research shows that incorporating client data for parameter initialization can be advantageous for federated learning frameworks \cite{yang2023greedy}. These aspects should be further investigated.

\textbf{Extremely Skewed Data Distribution}: In this work, we consider IID and non-IID data, implicitly assuming and requiring that  each client has all the class data points. However, this assumption does not always hold true in practical. For instance, some clients may have only data from one class or a few data points (e.g., a single point) from other classes. In this case, traditional supervised machine learning does not know how to build a classifier on this sparse data.
How to solve the problem of lack of class data and more skewed data distribution in the federated framework is a more challenging task. We leave this for public discussion. 

Furthermore, it is important to emphasize that although our framework is based on logistic regression, it has the adaptability to include other traditional machine learning methods, such as federated random forests \cite{liu2020federated}.
Nonetheless, it's crucial to acknowledge that these alternative models may not provide the same degree of interpretability as FLR, although they may show substantial enhancements in AUC and accuracy.


\section{Conclusion}
\label{sec:conclusion}
In this paper, our primary goal is to effectively analyze financial data distributed across clients or locations using machine learning algorithms, while also ensuring that the model possesses strong interpretability and explainability, under some practical concerns, such as distributed data across multiple locations, limited resources on a central server, and potential outliers generated by adversaries. 
To address these issues, we propose a federated logistic regression framework with robust aggregation methods, which can not only alleviate the effect of outliers, but also has very strong interpretability than complex models, such as neural networks and random forests. 
We evaluate our proposed framework on multiple public datasets, and the experimental results demonstrate that our proposed method can achieve comparable performance to classical centralized algorithms, such as LR, DT, and KNN on both binary and multi-class classification tasks. 
Furthermore, the results further demonstrate that regardless of IID and non-IID data, 1) our framework does not seem to be affected by the sample size $s$ much, especially when $s>=50$; 2) this also applies to the number of clients $M$. 
More importantly, our framework is robust to different percentages of outliers $p_{out}$. However, vanilla FLR with mean aggregation fails completely.
Additionally, in order to foster research in this area, we have made all datasets and the corresponding source code publicly available. 
Our next steps will revolve around addressing the issues and challenges discussed in Section \ref{sec:discussion}.

\bibliographystyle{elsarticle-num}
\bibliography{References}  

\clearpage
\appendix
\renewcommand{\thesection}{\Alph{section}}
\setcounter{section}{0}

\counterwithin{figure}{section}
\counterwithin{table}{section}

\section{Extra Results}
\label{sec:extra_results}
\subsection{Multicollinearity on CreditRisk}

\begin{figure*}[!htbp]
  \centering \captionsetup{hypcap=false} 
  \includegraphics[height=13.cm,keepaspectratio]{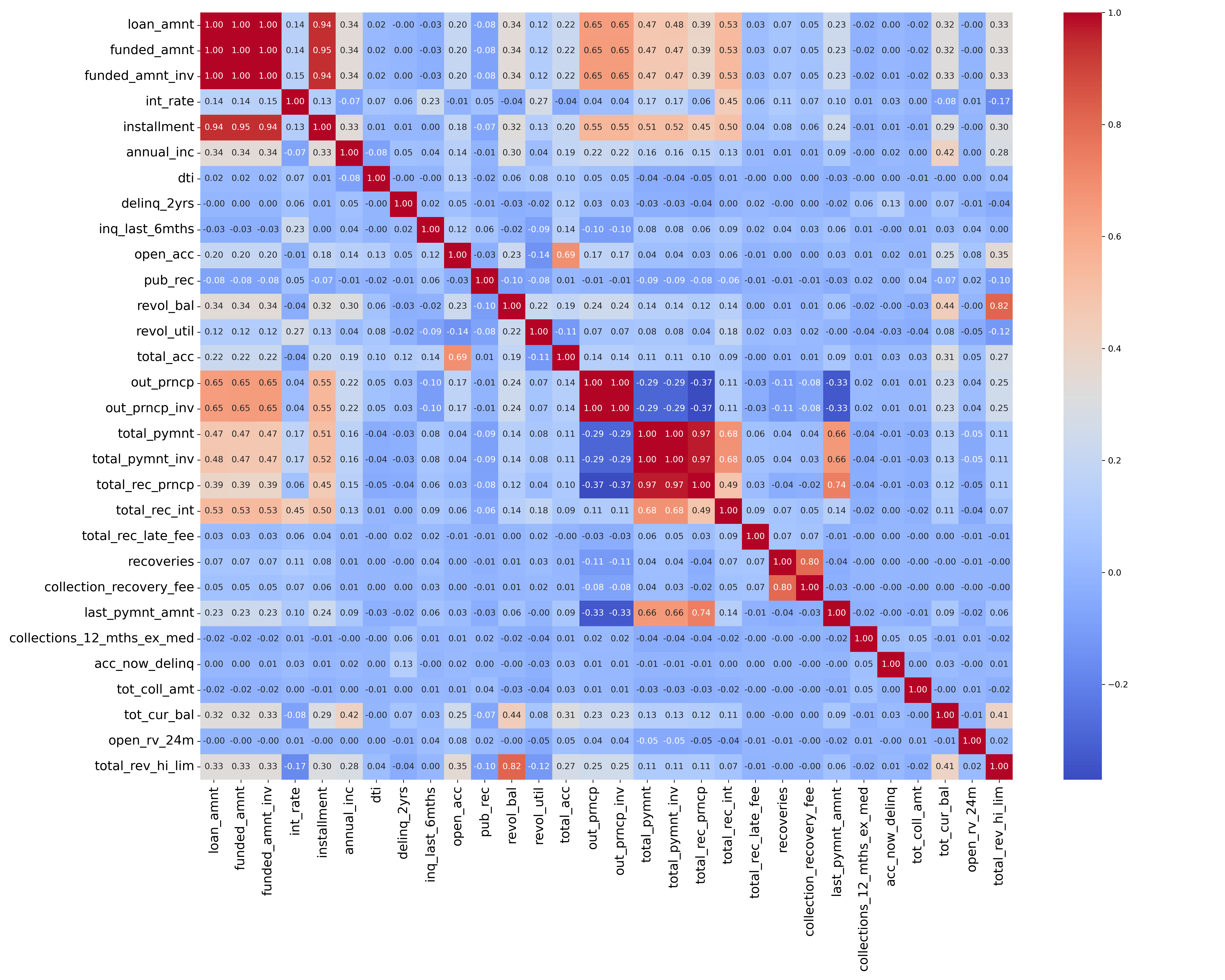}
  \caption{Feature correlation of CreditRisk before  removing any features.}
  \label{fig:correlation_before}
\end{figure*}

Figure \ref{fig:correlation_before} illustrates the presence of significant correlations in CreditRisk's original features before any features were removed.
For example, ``loan\_amnt'' and ``funded\_amnt'' are highly correlated with each other. 
Therefore, it is necessary to remove such correlations from the original features before performing subsequent analysis.

\subsection{Results on CreditScore with IID data}
\begin{figure}[htbp!]
    \centering
    \begin{minipage}{0.32\textwidth}
        \begin{center}
            \includegraphics[scale=0.33]{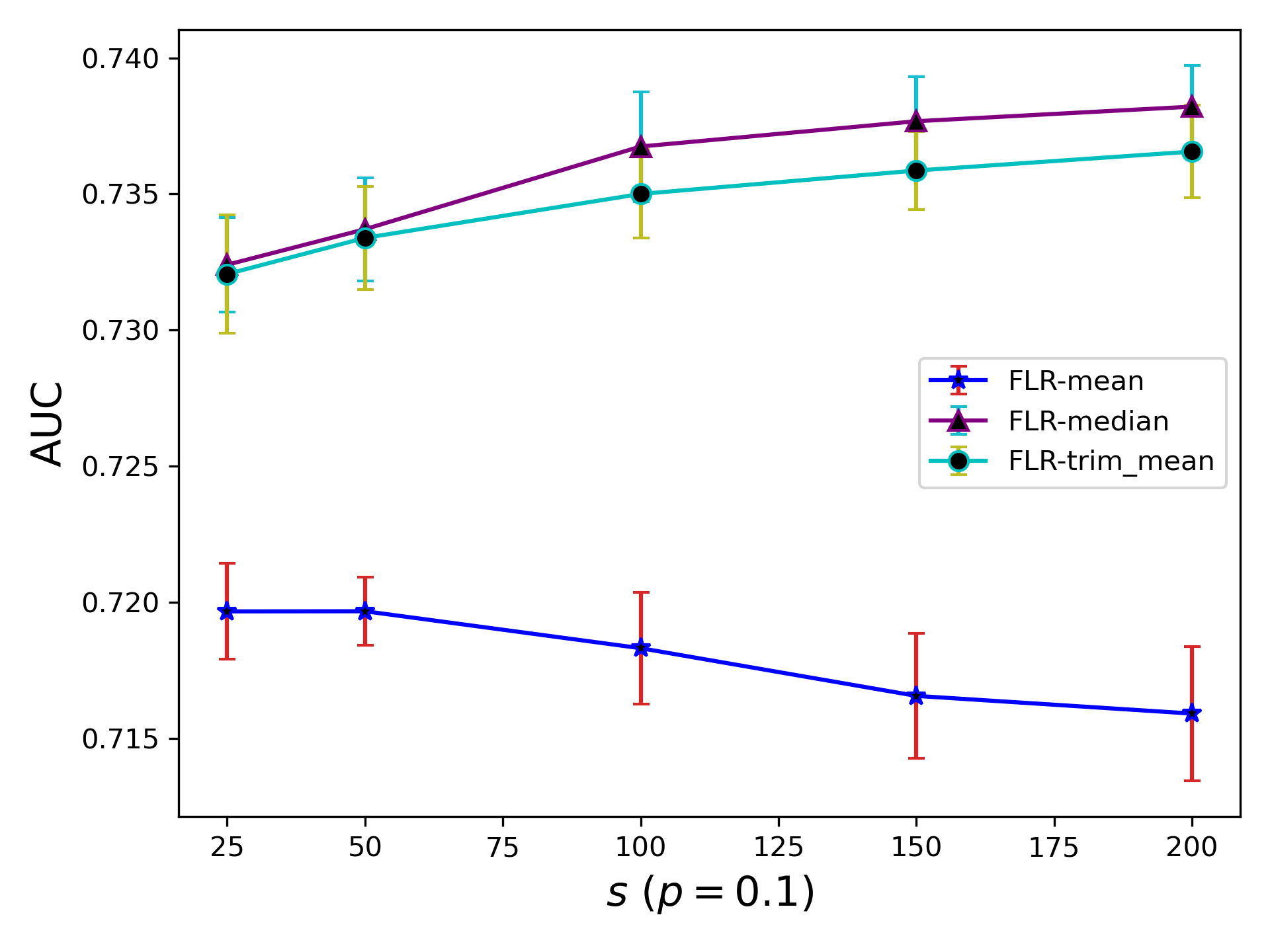}
        \subcaption{The effect of different sampling sizes $s$ with $M=100$ clients and $p_{out}=0.1$ on IID data.}
         \label{fig:creditscore_s_iid}
        \end{center}
    \end{minipage}\hfill
    \begin{minipage}{0.32\textwidth}
        \begin{center}
            \includegraphics[scale=0.33]{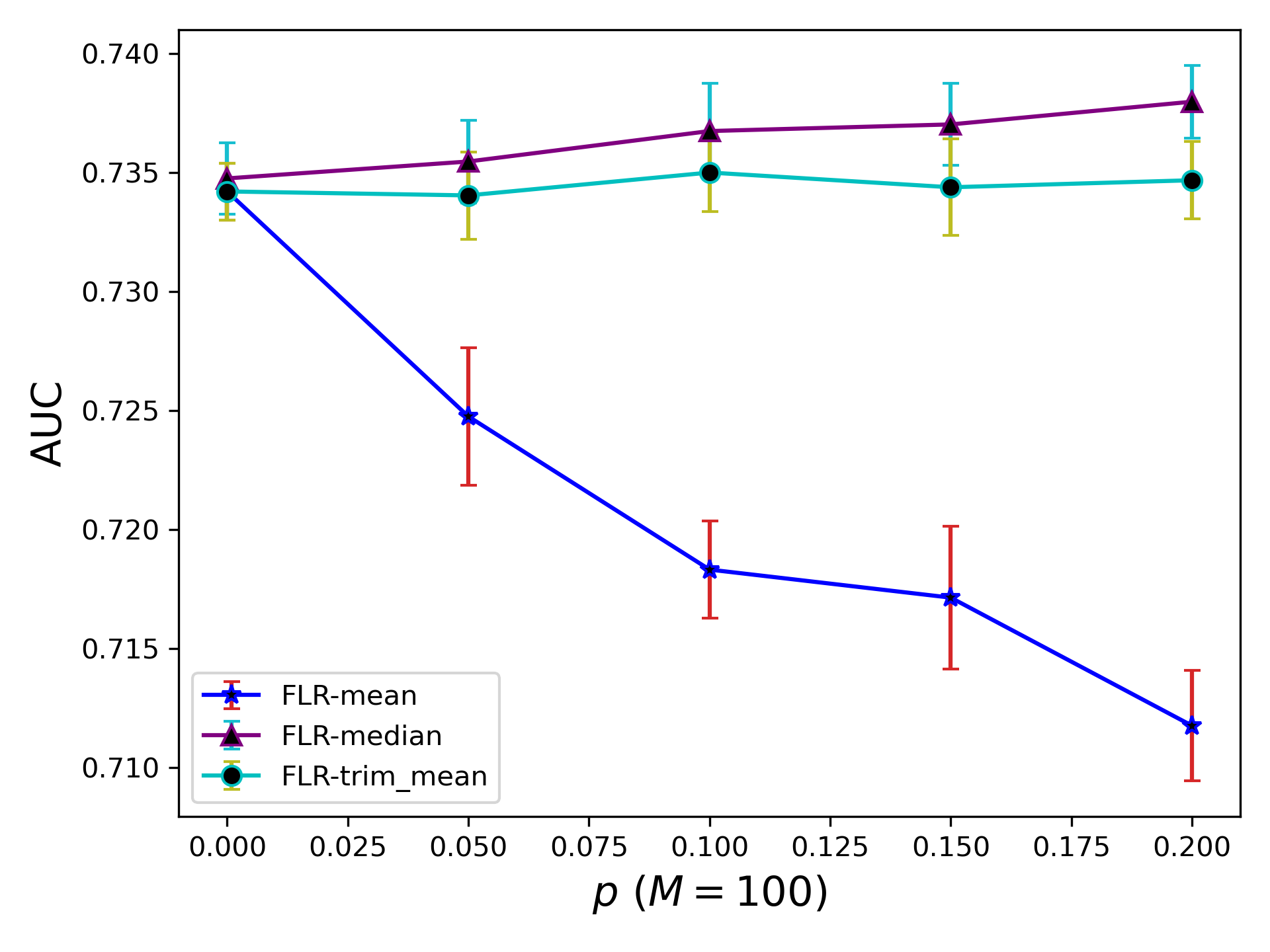}
       \subcaption{The effect of number of clients $M$ with fixed $p_{out}=0.1$ and $s=100$ on IID data.}
       \label{fig:creditscore_p_iid}
        \end{center}
    \end{minipage} \hfill
    \begin{minipage}{0.32\textwidth}
        \begin{center}
            \includegraphics[scale=0.33]{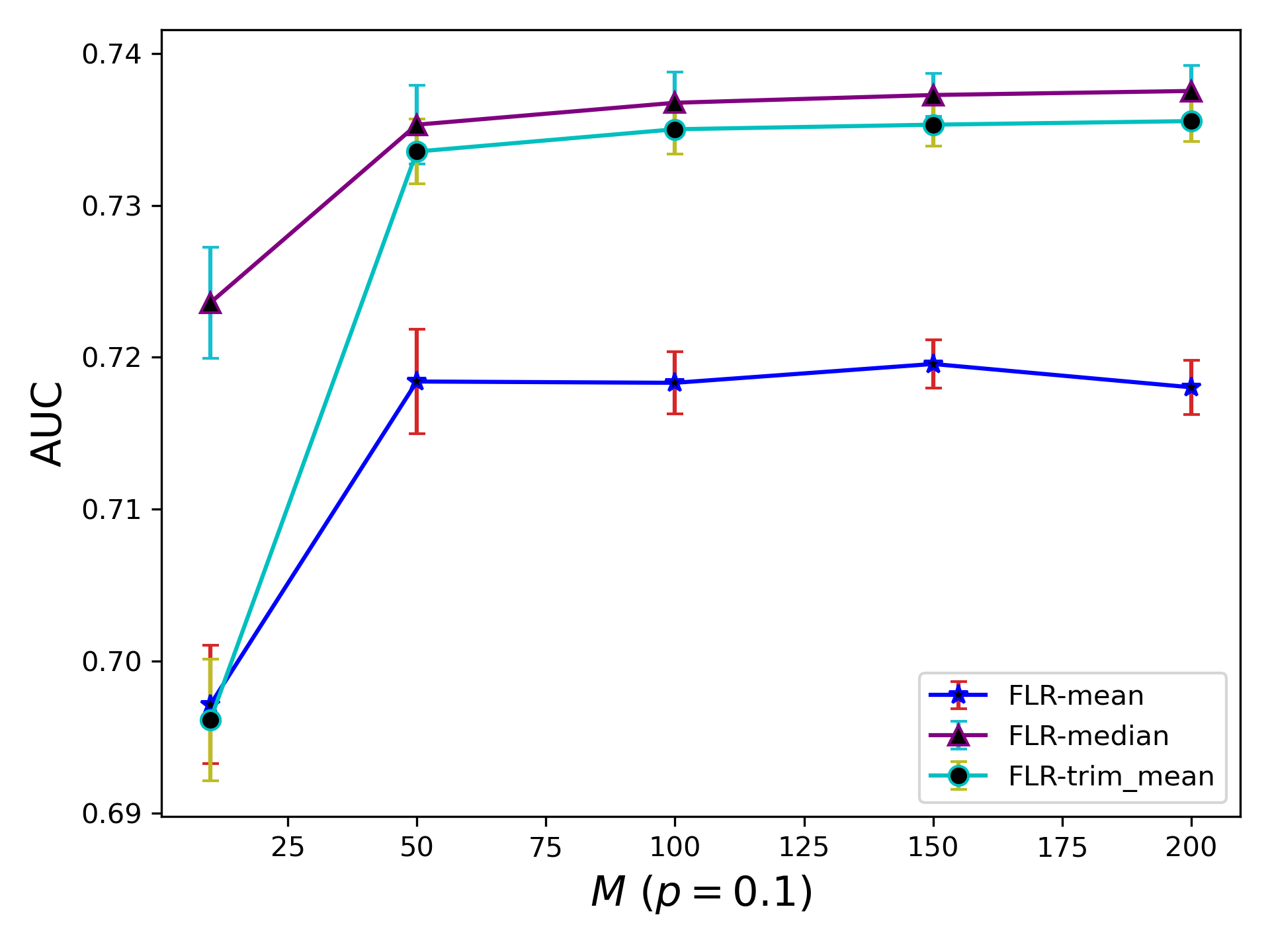}
       \subcaption{The effect of number of clients $M$ with fixed $p_{out}=0.1$ and $s=100$ on IID data.}
       \label{fig:creditscore_C_iid}
        \end{center}
    \end{minipage}
    \caption{Different cases on IID data.}
\end{figure}

Here, we reports all the results obtained by FLR with median aggregation on CreditScore with IID data, but no outliers, which demonstrates the similar performance as the results on non-IID data. 
Fig. \ref{fig:creditscore_s_iid} shows the effect of different sampling size $s$, Fig. \ref{fig:creditscore_p_iid} shows the effect of different percentages of outliers $p$, and Fig. \ref{fig:creditscore_C_iid} shows the effect of different number of clients $M$. 



\subsection{Feature Importance On CreditScore with IID data, but no outliers}
\begin{figure}[htbp!]
    \centering
    \begin{minipage}{0.33\textwidth}
        \begin{center}
            \includegraphics[scale=0.33]{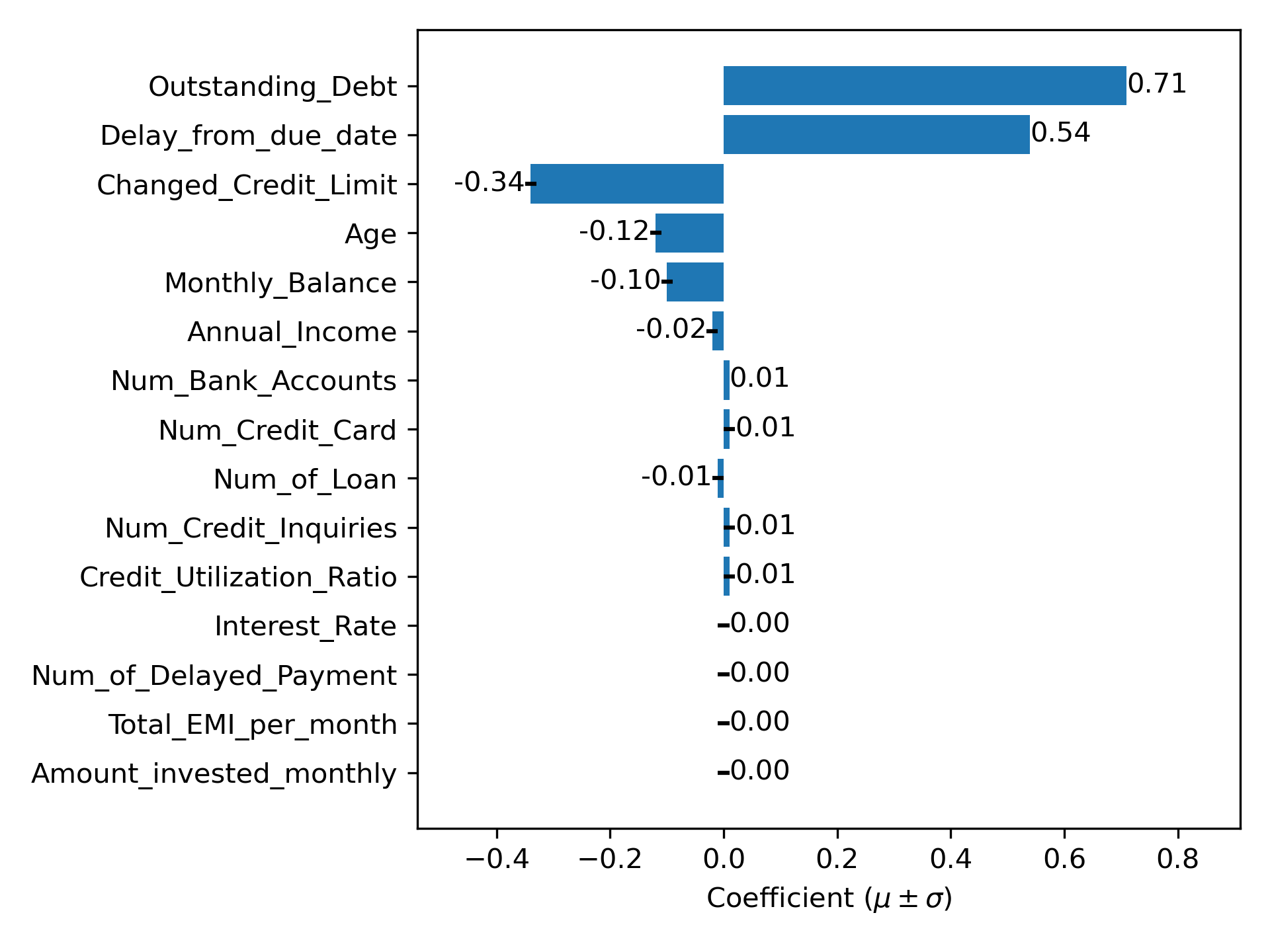}
        \subcaption{Poor vs. Rest}
        \end{center}
    \end{minipage}\hfill
    \begin{minipage}{0.33\textwidth}
        \begin{center}
            \includegraphics[scale=0.33]{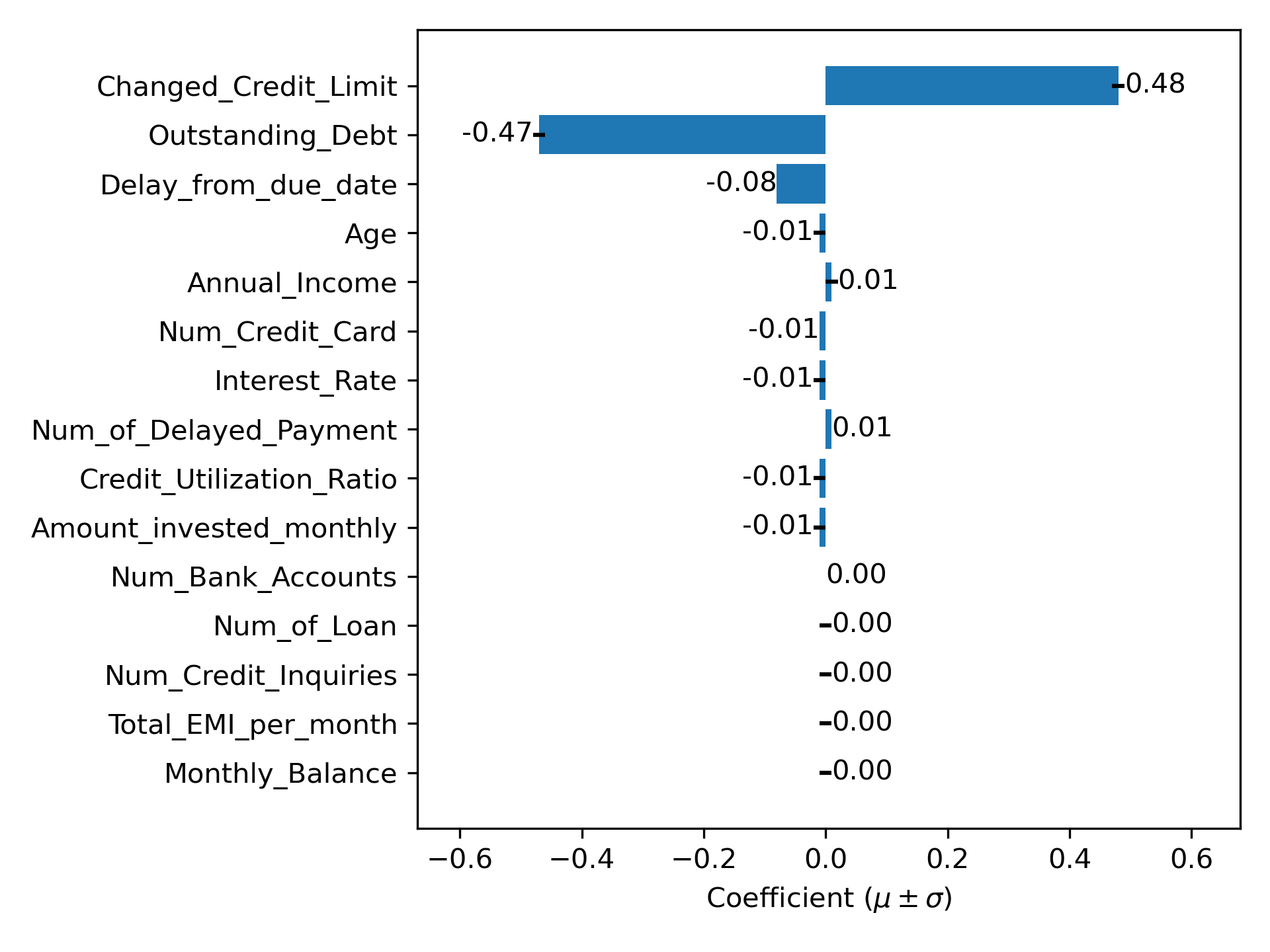}
       \subcaption{Standard vs. Rest}
        \end{center}
    \end{minipage} \hfill
    \begin{minipage}{0.33\textwidth}
        \begin{center}
            \includegraphics[scale=0.33]{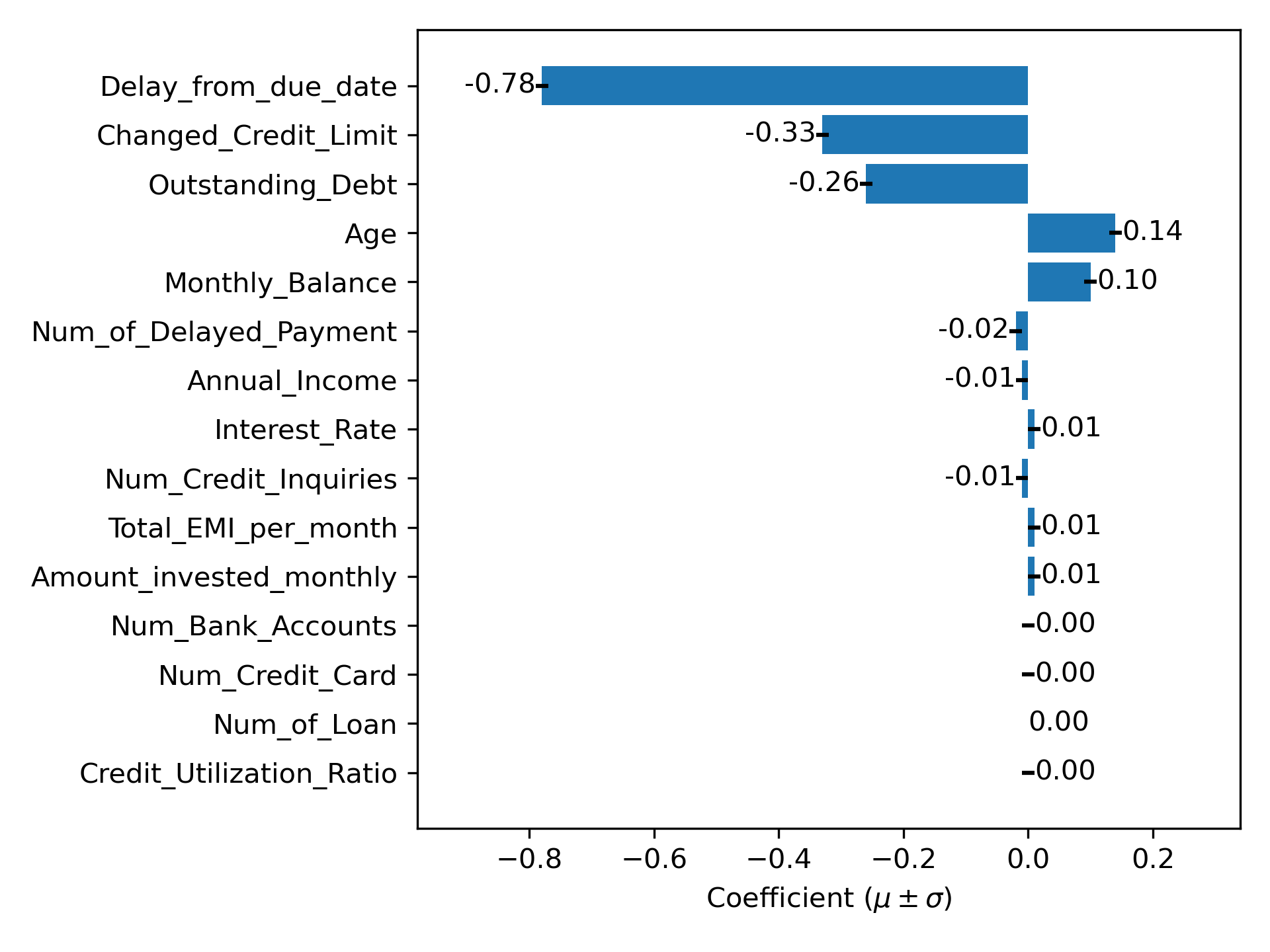}
       \subcaption{Good vs. Rest}
        \end{center}
    \end{minipage}
    \caption{Feature importances obtained by FLR with median aggregation on CreditScore with IID data, but no outliers.}
    \label{fig:feature_importance_iid_no_outliers}
\end{figure}

Fig. \ref{fig:feature_importance_iid_no_outliers} reports the feature importances obtained by FLR with median aggregation on CreditScore with IID data, but no outliers, which demonstrates the similar performance as the results on non-IID data.

\subsection{Results on CreditRisk}
\begin{figure*}[htbp!]
    \centering
    \begin{minipage}{0.5\textwidth}
        \begin{center}
            \includegraphics[scale=0.45]{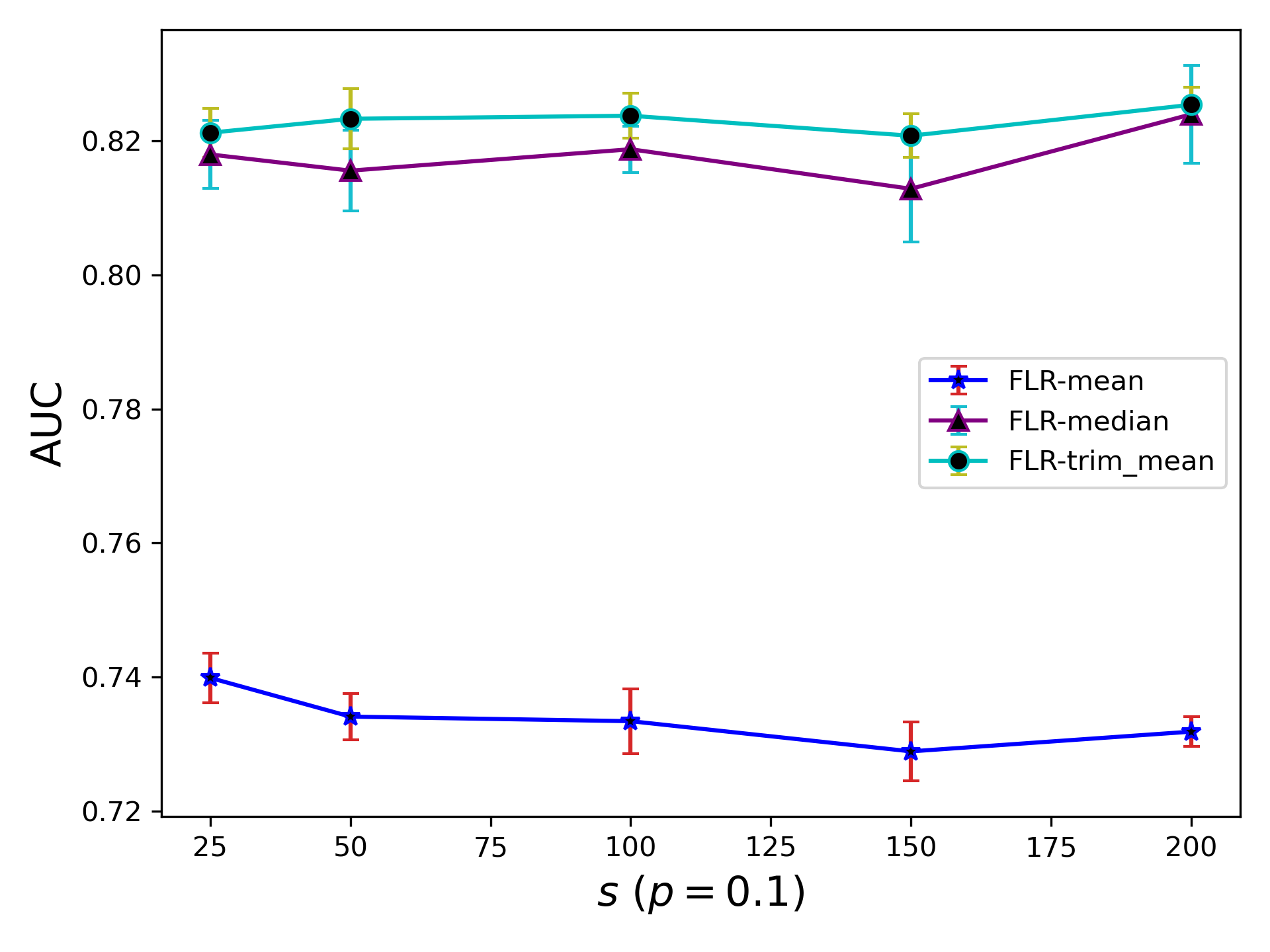}
       \subcaption{IID}
        \end{center}
    \end{minipage}\hfill
    \begin{minipage}{0.5\textwidth}
        \begin{center}
            \includegraphics[scale=0.45]{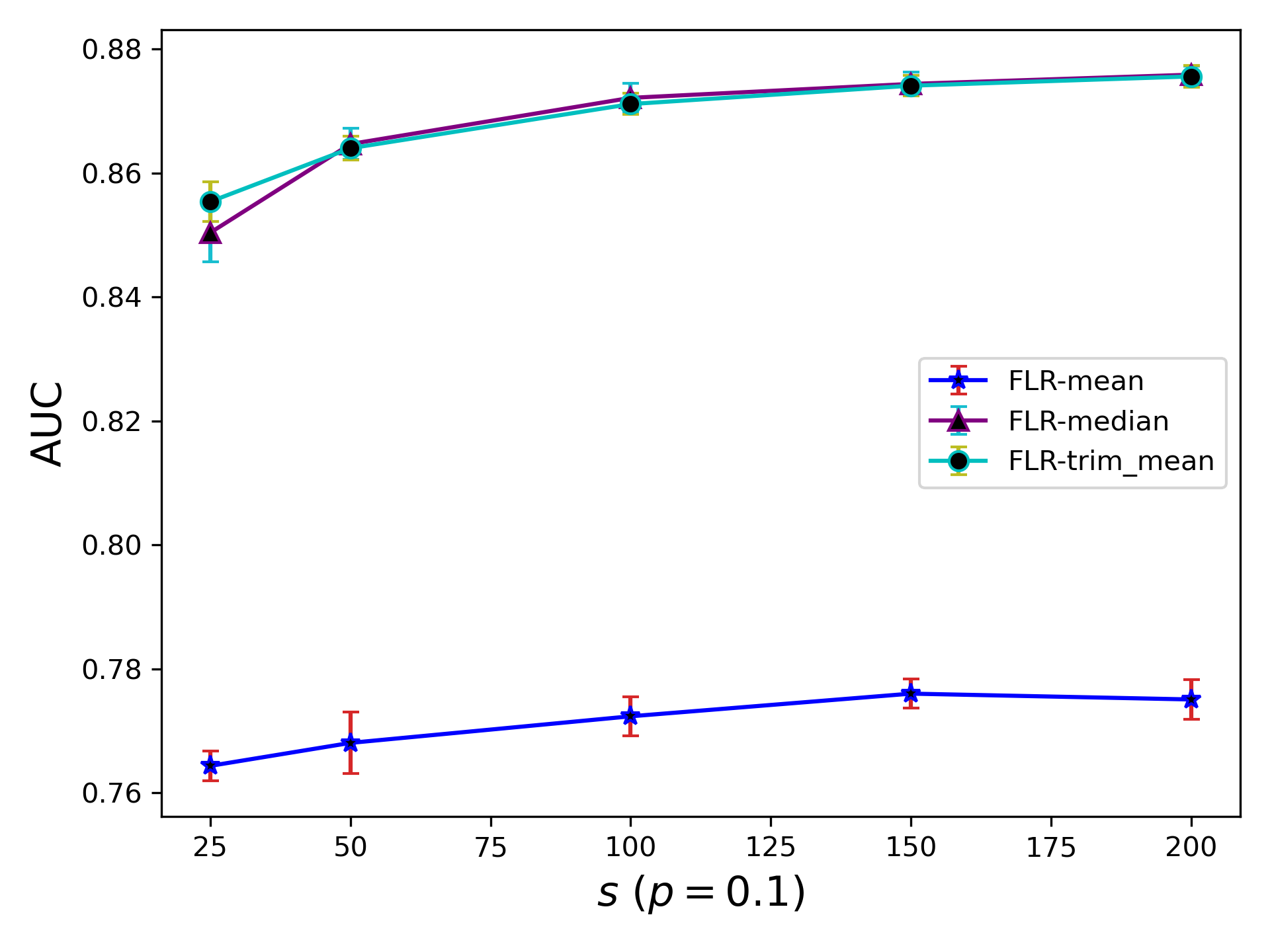}
       \subcaption{Non-IID}
        \end{center}
    \end{minipage}
    \caption{The effect of various sampling sizes $s$ on CreditRisk.}
    \label{fig:creditrisk_s}
\end{figure*}

\begin{figure}[!htbp]
    \centering
    \begin{minipage}{0.5\textwidth}
        \begin{center}
            \includegraphics[scale=0.45]{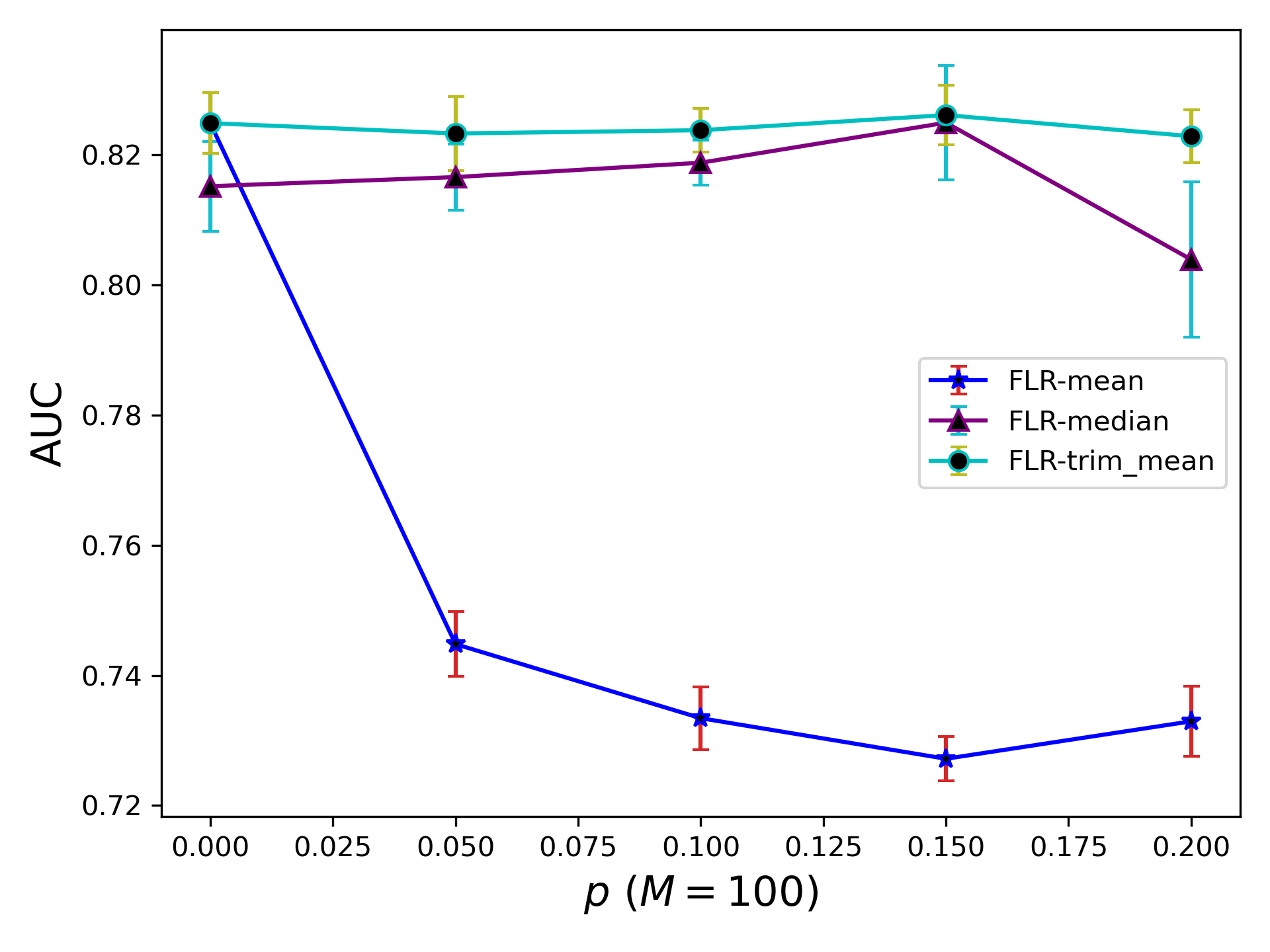}
       \subcaption{IID}
        \end{center}
    \end{minipage}\hfill
    \begin{minipage}{0.5\textwidth}
        \begin{center}
            \includegraphics[scale=0.45]{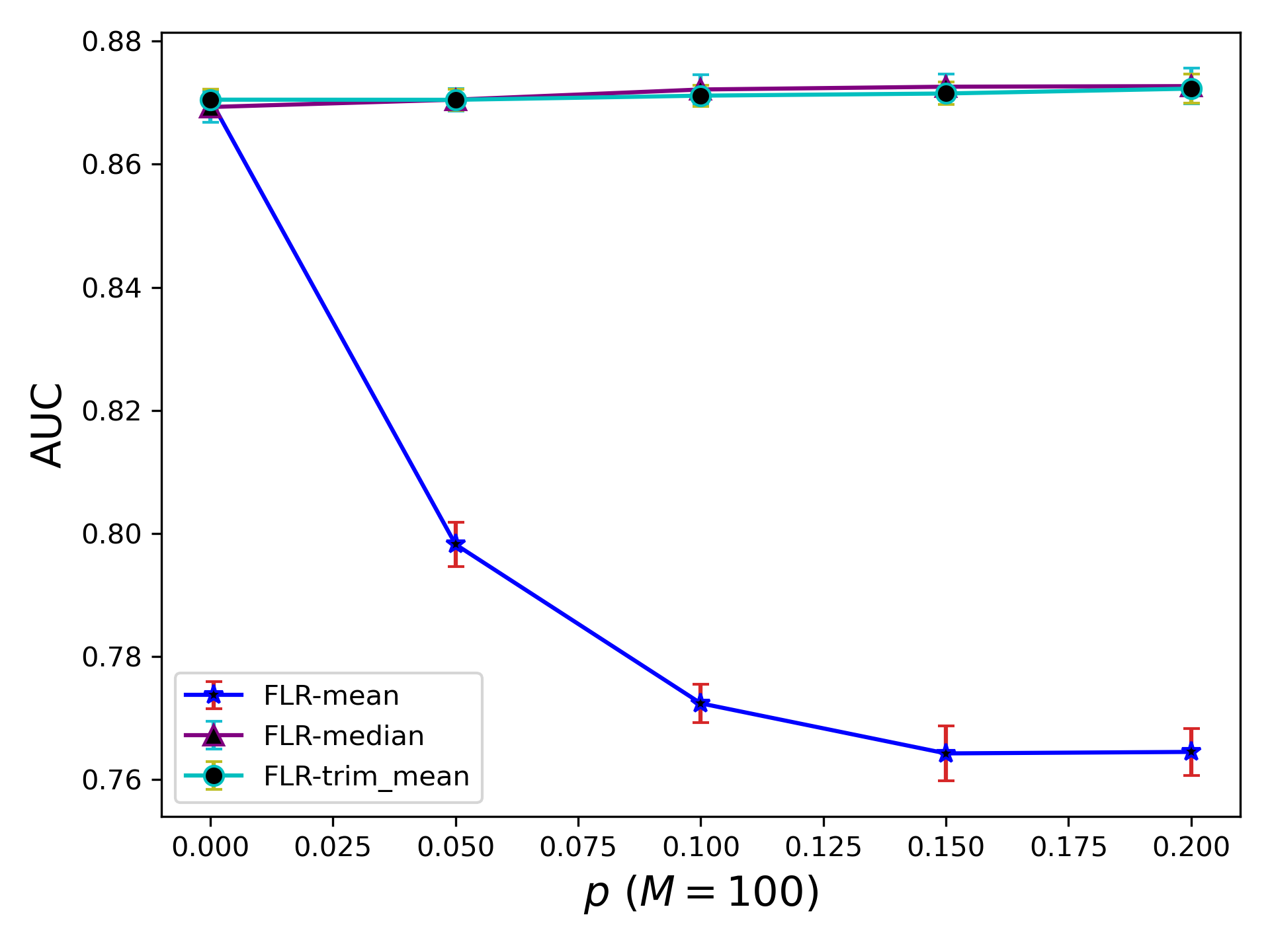}
       \subcaption{Non-IID}
        \end{center}
    \end{minipage}
    \caption{The effect of various percentages of outliers $p$ on CreditRisk.}
    \label{fig:creditrisk_p}
\end{figure}

\begin{figure}[!htbp]
    \centering
    \begin{minipage}{0.5\textwidth}
        \begin{center}
            \includegraphics[scale=0.45]{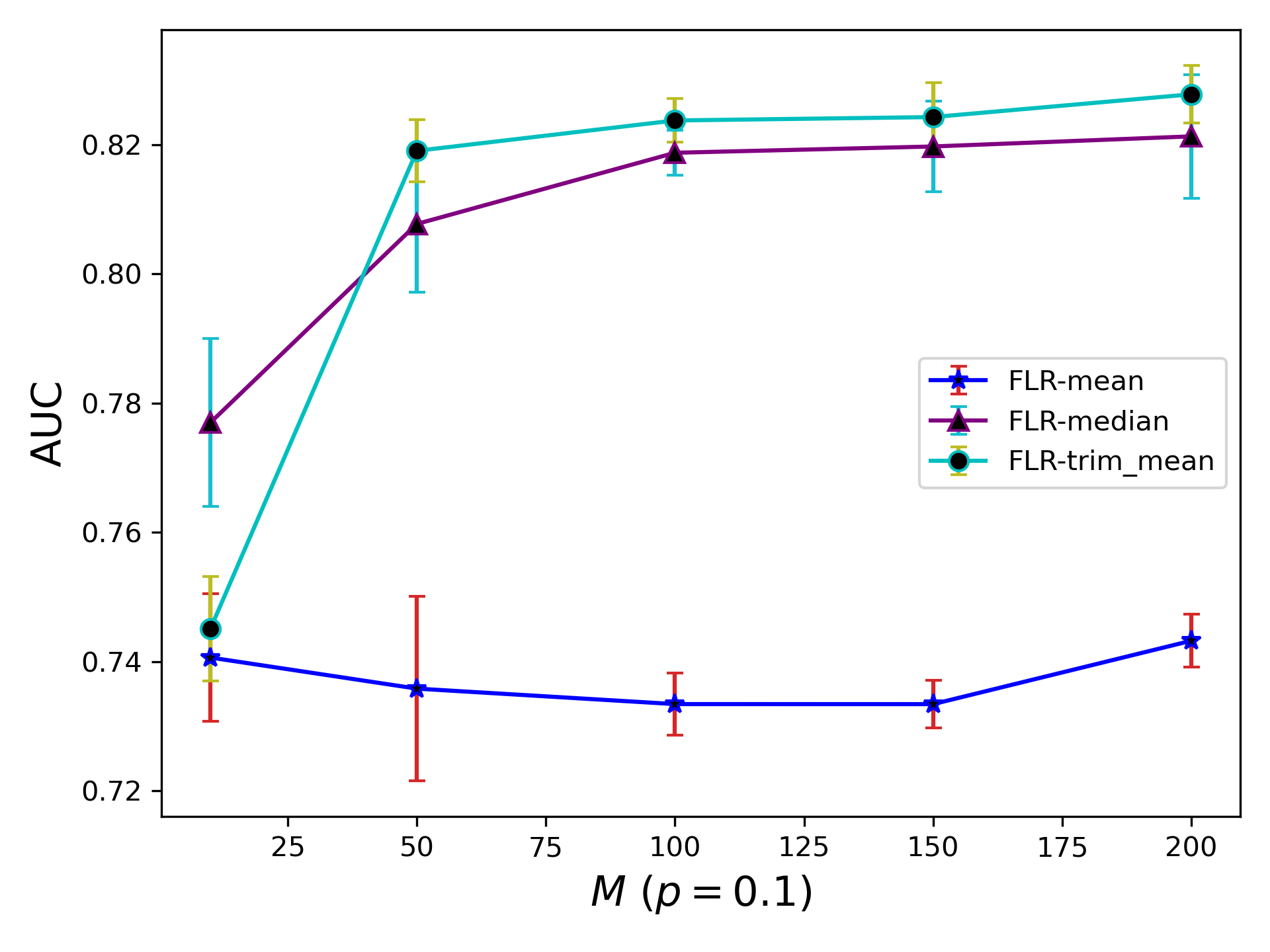}
       \subcaption{IID}
        \end{center}
    \end{minipage}\hfill
    \begin{minipage}{0.5\textwidth}
        \begin{center}
            \includegraphics[scale=0.45]{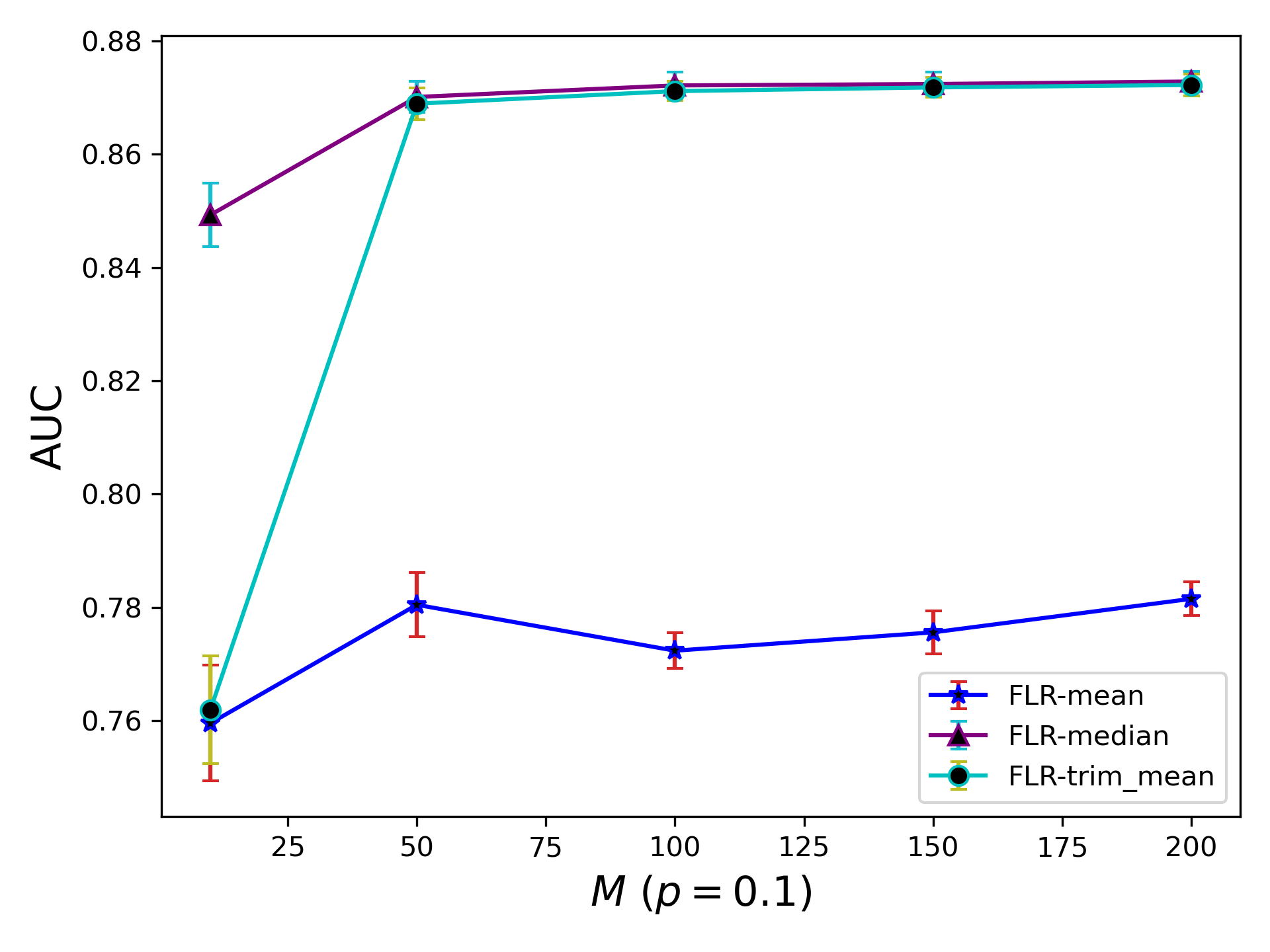}
       \subcaption{Non-IID}
        \end{center}
    \end{minipage}
    \caption{The effect of number of clients $M$ on CreditRisk.}
    \label{fig:creditrisk_C}
\end{figure}

Here, we reports all the results obtained by FLR with median aggregation on CreditRisk, but no outliers,  which demonstrates
the similar performance as the results on CreditScore. 
Fig. \ref{fig:creditrisk_s} shows the effect of different sampling size $s$, Fig. \ref{fig:creditrisk_p} shows the effect of different percentages of outliers $p$, and Fig. \ref{fig:creditrisk_C} shows the effect of different number of clients $M$. 

\clearpage
\vspace{-1cm}

\end{document}